\documentclass[manuscript]{acmart} 
\usepackage[nolist, nohyperlinks]{acronym}
\usepackage{booktabs}

\begin{acronym}
    \acro{GEM}{Generalised Expectation-Maximisation}
\end{acronym}

\copyrightyear{2020}
\acmYear{2020}
\setcopyright{acmlicensed}\acmConference[RecSys '20]{Fourteenth ACM Conference on Recommender Systems}{September 22--26, 2020}{Virtual Event, Brazil}
\acmBooktitle{Fourteenth ACM Conference on Recommender Systems (RecSys '20), September 22--26, 2020, Virtual Event, Brazil}
\acmPrice{15.00}
\acmDOI{10.1145/3383313.3412253}
\acmISBN{978-1-4503-7583-2/20/09}

\begin{document}

\title[Making Neural Networks Interpretable with Attribution]{Making Neural Networks Interpretable with Attribution: Application to Implicit Signals Prediction}

\author{Darius Afchar}
\author{Romain Hennequin}
\affiliation{%
  \institution{Deezer Research}
  \city{Paris}
  \country{France}
}
\email{research@deezer.com}

\begin{abstract}
Explaining recommendations enables users to understand whether recommended items are relevant to their needs and has been shown to increase their trust in the system. More generally, if designing explainable machine learning models is key to check the sanity and robustness of a decision process and improve their efficiency, it however remains a challenge for complex architectures, especially deep neural networks that are often deemed "black-box". In this paper, we propose a novel formulation of interpretable deep neural networks for the attribution task. Differently to popular post-hoc methods, our approach is interpretable by design. Using masked weights, hidden features can be deeply attributed, split into several input-restricted sub-networks and trained as a boosted mixture of experts. Experimental results on synthetic data and real-world recommendation tasks demonstrate that our method enables to build models achieving close predictive performances to their non-interpretable counterparts, while providing informative attribution interpretations.
\end{abstract}



\maketitle

\section{Introduction}

In recent years, deep neural networks have been successfully used in a wide range of fields to predict, classify, recommend and generate content, often achieving state-of-the-art results. However, understanding the behaviour of such modern machine learning models remains a challenge compared to simpler methods with transparent computation processes (\textit{e.g.} linear regressions or decision trees). 

Yet, \textit{explainability} is needed in many fields where the sanity, robustness and fairness of a decision has to be checked, for instance in the medical field, autonomous driving or business analytics, hampering the adoption of those highly-performing models. In the field of recommender systems, although providing explanations is not a necessity, it has been widely shown to improve users satisfaction and trust in the system~\cite{herlocker2000explaining, sinha2002role, tintarev2007survey}.
The use of neural networks does not consistently lead to an improvement in recommendation~\cite{dacrema2019we}, but their ubiquity across various recent methods~\cite{shenbin2020recvae, hidasi2015session, li2017collaborative, he2017neural, liang2018variational, he2018nais, van2013deep} brings us to the study of \textbf{neural networks interpretability}.   

In particular, real-world applications often come with the hardship of taking into account \textbf{implicit signals}~\cite{hu2008collaborative} from users, namely signals with a broad semantic, that indirectly reflect inaccessible, high-level or conceptual data, as personal tastes of users or specific contexts.
Neural networks can provide a way to model the complex, sometimes multimodal, nature of such implicit signals. As an example, for the implicit \textit{sequential skip prediction} challenge in music streaming sessions\footnote{\label{wsdm} See \textit{WSDM Skip Prediction Challenge}: \url{https://aicrowd-design.netlify.app/template-challenge-overview}}, traditional models have been outmatched in favour of deep-network-based approaches~\cite{zhu2019session, hansen2019modelling, chang2019sequential}. Interpretation was not particularly studied is the latter works, however, we argue that producing interpretations is beneficial to our understanding of the studied implicit signals, which may allow to wittily leverage implicit users feedback in recommender systems by making them more explicit.

In this paper, we study the interpretation of implicit signals through the lens of \textbf{feature attribution}: the behaviour of a model is simplified to the knowledge of the input dimensions that are primarily used to make a prediction. Indeed, attribution is relevant for implicit signals to unveil their underlying nature. In the example of \textit{skip prediction}, it enables to discriminate between the case of a user disliking a song because of its music content, and another exploring the catalog and thus quickly skipping through content, which should be interpreted as distinct feedbacks by a recommender system. This simple dichotomy is crucial for music streaming services to refine user music profiles and is currently underexploited due to the implicitness of \textit{skips}. If attribution is straightforward with linear models, it is more difficult to trace the origin of a prediction through usually multi-layered neural networks.

We propose the formulation of a novel class of deep neural networks that are intrinsically interpretable in term of feature attribution. In detail, by using mask constraints in linear layers, we define a deep structured neural network that allows to trace for every neuron what input data its computation is based on. This allows to emulate several expert sub-networks, each based on a specific restriction of the input. Experts are constrained to be residuals of simpler available experts in order to enforce sparsity. This mixture of expert is then jointly trained using a \textit{\ac{GEM}} algorithm. For inference, our network both produces a prediction and an attribution estimation. This method can be applied to make many modern deep architectures interpretable (\textit{e.g.} Transformer).

Our contributions are the following:
\begin{itemize}
    \item We formulate a way to make deep neural networks architecture interpretable while achieving almost similar performances as their non-interpretable counterpart for several recommendation tasks ;
    \item we derive a fast joint training algorithm for this novel architecture inspired by \textit{boosting} ;
    \item we demonstrate the effectiveness of our model on the prediction and interpretation of implicit signals and its application for the real-world task of \textit{sequential skip prediction}.
\end{itemize}

If our method was designed with implicit signals and recommendation in mind, it is not limited to them and could be applied to a broad class of attribution tasks. It should also be noted that our intrinsically interpretable method does not preclude the use of popular post-hoc methods and both approaches can complement each other.

\section{Related work}

\label{sec:2}

\subsection{Interpretability for deep neural networks}

Interpretability in machine learning is an expanding research field that encompasses many different methods.
A popular branch of interpretability aims at providing a post-hoc analysis on how the output of a model is related to its input data. This is the case of the \textit{LIME} method~\cite{ribeiro2016should}, that locally computes a linear model of a trained black-box model, thus providing a simplified explanation of how each input dimension influences the predicted target label for a given input space region. With the same idea, \textit{DeepRED}~\cite{zilke2016deepred} uses decision trees as the simplified proxy model, allowing to interpret a deep model as a composition of extracted rules. Going further, methods such as \textit{DeepLIFT}~\cite{shrikumar2017learning}, \textit{LRP}~\cite{bach2015pixel} and other saliency methods~\cite{simonyan2013deep, zhou2016learning, selvaraju2017grad, smilkov2017smoothgrad} or game-theory-based \textit{SHAP}~\cite{NIPS2017_7062} can interestingly propagate feature importance values throughout the layers of a deep model, yielding interpretability up to a neuron-wise granularity.
Another branch of interpretability is focused on the elaboration of explanation-producing models. This can be done in a supervised manner as in \cite{hendricks2016generating, kim2018interpretability} when data about the desired output explanation are available and well-posed, allowing to produce high-level explanations that are more human understandable, or in an unsupervised manner using intrinsically interpretable model.
\textit{Intrinsic} interpretability is a desirable property for high stakes decision models~\cite{rudin2019stop}, but also for researchers to inspect, understand and improve how neural network components manipulate data. This is for instance the case of the attention mechanism and its extended multi-head attention module~\cite{vaswani2017attention}, widely used in natural language processing tasks, that reveals the specialisation of heads into different classes of reading mechanisms for words in a sentence~\cite{voita2019analyzing}, or atoms in a molecules to form chemical patterns~\cite{maziarka2020molecule}.
Using information theory, \textit{InfoGAN}~\cite{chen2016infogan} learns to disentangle latent representations during training, making them interpretable and manipulable. The \textit{information bottleneck} principle~\cite{tishby2000information} is also a promising concept for interpretability that was successfully used in \cite{schulz2020restricting} for feature attribution.
Using this taxonomy, our method is an intrinsic interpretation method focused on the attribution problem. We do not assume to have access to target explanations, making the interpretability task unsupervised. Differently from information-theoretic and variational methods, we do not require priors on the attributions, allowing to solve a broader class of attribution problems (see section \ref{objective}). We additionally leverage the natural interpretation power of multi-head attention modules in our chosen deep models, though not being limited to it.


\subsection{Selection}

Our attribution method is related to \textit{generalised additive models}~\cite{hastie1990generalized} that model a function as a sum of univariate sub-functions. This formulation is intrinsically interpretable as the contribution of each input feature can be assessed by inspecting corresponding univariate functions. Going further, pairwise interactions can be added, as in \cite{lou2013accurate}: freezing trained univariate functions, the authors add bivariate functions that are trained on residual points in a boosting-like manner~\cite{freund1999short}. In this spirit, our method extends \cite{lou2013accurate} to any multivariate functions. However, residuality is replaced by general gating functions on classification confidence of child functions with fewer input variables. This formulation is closely related to the \textit{mixture of experts}~\cite{jacobs1991adaptive, jordan1994hierarchical}, allowing us to train our model jointly instead of iteratively as in \cite{lou2013accurate}.
Our formulation of ensemble learning is also reminiscent of \textit{subset selection}~\cite{friedman2001elements}. To avoid the high combinatorial number of best subset candidates, we restrict their space to a reasonable cardinality using human knowledge. 

\subsection{Structured networks}

We use a judiciously structured deep neural network to emulate several deep sub-networks acting as our different multivariate experts.
Our original inspiration comes from \textit{YOLO}~\cite{redmon2016you}, a paramount model in the object detection literature. Interestingly, the network outputs several candidates bounding boxes and self-confidence scores, only one predictor is then selected, leading to a specialisation of predictors to specific classes of objects, as reported by the authors. In \cite{voita2019analyzing}, the authors also report a natural specialisation of different components of a multi-attention module. Our method aims at inducing this specialisation to predefined input subsets of interest for interpretability.
We manipulate and route neurons by blocks, which can be related to \textit{capsule networks}~\cite{hinton2011transforming}. Cunningly structuring a neural network has indeed been demonstrated to produce intrinsic interpretability, as in the recent \textit{RPGAN}~\cite{voynov2019rpgan}. However, the routing process is fixed in our method as interpretation subsets are hyper-parameters in this work. We have explored the use of dynamic subsets, which draws our structured network architecture closer to the latter methods, but leave it out of the scope of this paper. 


\section{Proposed Method}

\label{sec:3}

\begin{figure}[h]
  \centering
  \includegraphics[width=\linewidth]{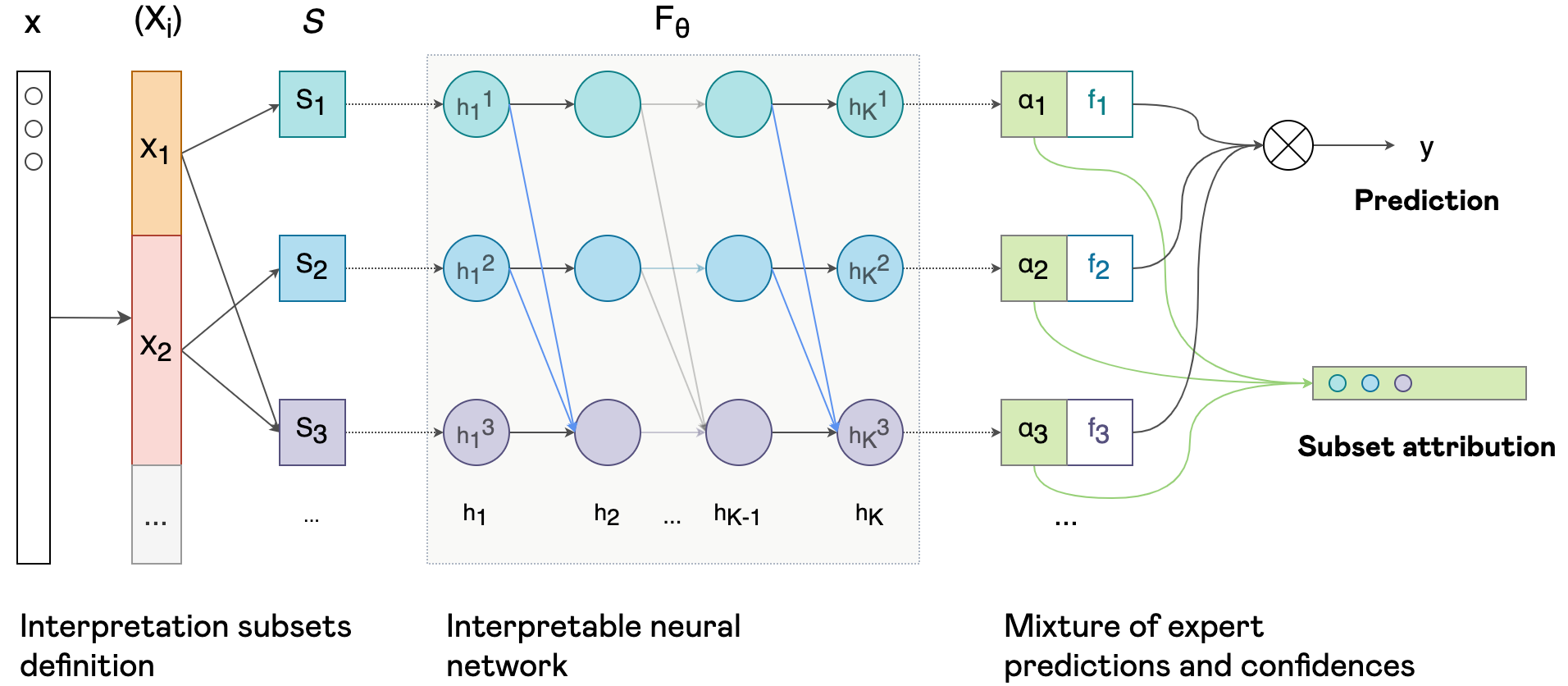}
  \caption{\textit{\textbf{Method overview}} An input $x$ is partitioned into coherent groups of features $\{X_1, ... X_N \}$, combined into subsets of interest for interpretation $\{S_1, ... S_H \}$, then fed to a structured neural network that preserves subset dependence, the output of which is combined in a mixture-of-expert manner to provide a prediction $y$. Additionally, sub-networks specialised on each interpretation subset can be accessed to provide an attribution vector to interpret the origin of the prediction.} 
  \label{overview}
\end{figure}

In this section, we introduce the different building blocks of our method. An overview is given in figure~\ref{overview}.

\subsection{Problem formulation}
\label{objective}

We consider the \textbf{supervised classification} setting where $x \sim X \in \mathbb{R}^n$ is the input random variable and $y \sim Y \in \{0,1\}$ its target label, our model $f_\theta: x \mapsto y$ maps the sample of $X$ to $Y$ and is parametrised by $\theta$. In this paper, we restrict our study to the \textbf{binary case} for $Y$, but our method can be extended to the multi-class and continuous case with little effort, but because the latter settings need additional discussion and experiments, we leave them for future work. Parallel to the classification task, we want our model to solve an \textbf{attribution} task by yielding interpretation masks for its inputs that highlight the features that were the most relevant to make the prediction. We introduce a random mask variable $m \sim M$ that takes values in a finite space $\mathcal{M} \subset \{0,1\}^n$ and depends on $X$ such that: we want \textbf{\textit{(a)} the model to be able to accurately predict $Y$ from $X \circ M$}, where $\circ$ denotes the Hadamard product, while \textbf{\textit{(b)} having $M$ as sparse as possible}. We do not assume $M$ to be observable, making it a latent variable for our model.

The way we choose to solve the attribution problem is by considering different restrictions of the input $X \circ M$, and feeding them into several sub-models $f^m_\theta$ (experts). We then average expert predictions by introducing several associated selection functions $\alpha^m_\theta$ that activate different experts depending on the input:
\begin{equation}
    \label{mixture}
    f_\theta(x) = \sum_{m \in \mathcal{M}} \alpha^m_\theta(x \circ m) f^m_\theta(x \circ m) / \sum_{m \in \mathcal{M}} \alpha^m_\theta(x \circ m)
\end{equation}
This latter ensemble technique is closely related to the \textit{mixture of experts}~\cite{jacobs1991adaptive, jordan1994hierarchical},
but with input restrictions.

Our objective is to \textit{\textbf{(a)}} find a maximum solution for the likelihood $p_\theta(y|x) = \sum_{m \in \mathcal{M}} p_\theta(y | x, m)p_\theta(m|x)$, and \textit{\textbf{(b)}} ensure the sparsity of $m$ by design. As detailed in section \ref{sec:training}, we will maximise the likelihood using \ac{GEM}~\cite{bishop2006pattern}, which involves the incremental update of the following conditional expectation $\mathcal{Q}$:

\begin{equation}
    \mathcal{Q}(\theta, \theta_{\textrm{old}}) = \mathbb{E}_{x\sim X, y \sim Y}[\sum_{m \in \mathcal{M}} p_{\theta_{\textrm{old}}}(m | x) \ln p_\theta(y | x, m) ]
\end{equation}
The posterior distribution $p_\theta(M | X)$ is modelled by the selection functions $\alpha^m_\theta$. Depending on the problem, the marginal likelihood $p_\theta(y | x, m)$ has to be modelled in different ways, we derive $\mathcal{Q}$ in the binary case with standard assumptions~\cite{bishop2006pattern}:
\begin{align}
    p_\theta(y | x,m) &\sim \mathcal{B}(f^m_\theta(x \circ m)) \\
\label{eq:binary_derivation} \arg \max_\theta \mathcal{Q} & = \arg \min_\theta \mathbb{E} \sum_{m \in \mathcal{M}} \alpha^m_{\theta_{\textrm{old}}}(x \circ m) \textrm{BCE}(y,f^m_\theta(x \circ m))
\end{align}
where $\textrm{BCE}$ stands for \textit{Binary Cross-Entropy}. In the following sections we develop the different terms of equation \eqref{eq:binary_derivation}: in section \ref{sec:latent_red} we prune the mask candidates space $\mathcal{M}$, we then study the computation of $\alpha^m_\theta$ to enforce the sparsity of $M$ by design in section \ref{sec:selection} and finally we detail the architecture of $f^m_\theta$ in section \ref{sec:nn}.

\subsection{Latent space reduction}
\label{sec:latent_red}

Our formulation requires to sum likelihoods conditioned on the space $\mathcal{M}$ of all candidate interpretation masks. However, the number of mask is exponential with the $n$ dimensions of $X$, making the computation intractable for realistic values of $n$. The working hypothesis of this paper is that \textbf{we do not need to consider all possible masks for $M$}.

Our first approximation consists in considering that the masks are \textbf{group-sparse}~\cite{huang2010benefit}. Feature attribution can indeed lack robustness by yielding noisy or incoherent subset of features that act as adversarial solutions to the interpretability task and make them less human-understandable~\cite{alvarez2018robustness}. Only allowing group-sparsity of coherent subsets of features mitigates the effect by regularising the allowed solutions. We thus partition $X$ into $N$ disjoint subsets $\mathcal{X} = \{ X_1, \dots X_N \}$. In the example \textit{skip prediction} attribution task we have mentioned in the introduction, we could partition the input space into \textit{interaction features} ($X_1$) versus \textit{musical features} ($X_2$) to understand the origin of a \textit{skip} feedback.

In practical applications, we often know the \textbf{structure} of $M$ and the sparsity patterns we can obtain~\cite{huang2011learning, zhao2009composite}.
In such cases, \textbf{we can further prune the set of mask candidates to only match consistent patterns}.
We denote $\mathcal{S}$, the resulting subset of size $H$ of all possible combinations of subsets $X_i$: $\mathcal{S} = \{ S_1, \dots S_H \} \subset \mathcal{P}(\mathcal{X})$, with $\mathcal{P}$ the powerset. With the previous example of \textit{skip prediction}, if we had further split \textit{musical features} into \textit{genre} ($X_2$) and \textit{mood estimation} ($X_3$), it would be coherent to consider the subsets $\{X_2\}$, $\{X_3\}$, and the aggregated $\{X_2, X_3\}$ musical features.

Doing so, instead of summing on a space of size $2^n$ for $p(m|x)$, we assume we can work with a reasonable number $H \ll 2^n$ of masks-by-block candidates for interpretations. Of course, $\mathcal{X}$ and $S$ can be manually tuned to obtain coarser or finer level of interpretability. In the following sections, we denote $\mathcal{M} = \{ m_1, \dots m_H \}$ the interpretation masks we assume given and fixed and isomorphic to a given $\mathcal{S}$ through the relation $X \circ m_i = S_i$.

\subsection{Selection functions}
\label{sec:selection}
\subsubsection{Toy examples}
\label{sec:selection_dep}
\begin{figure}[h]
  \centering
  \includegraphics[width=\linewidth]{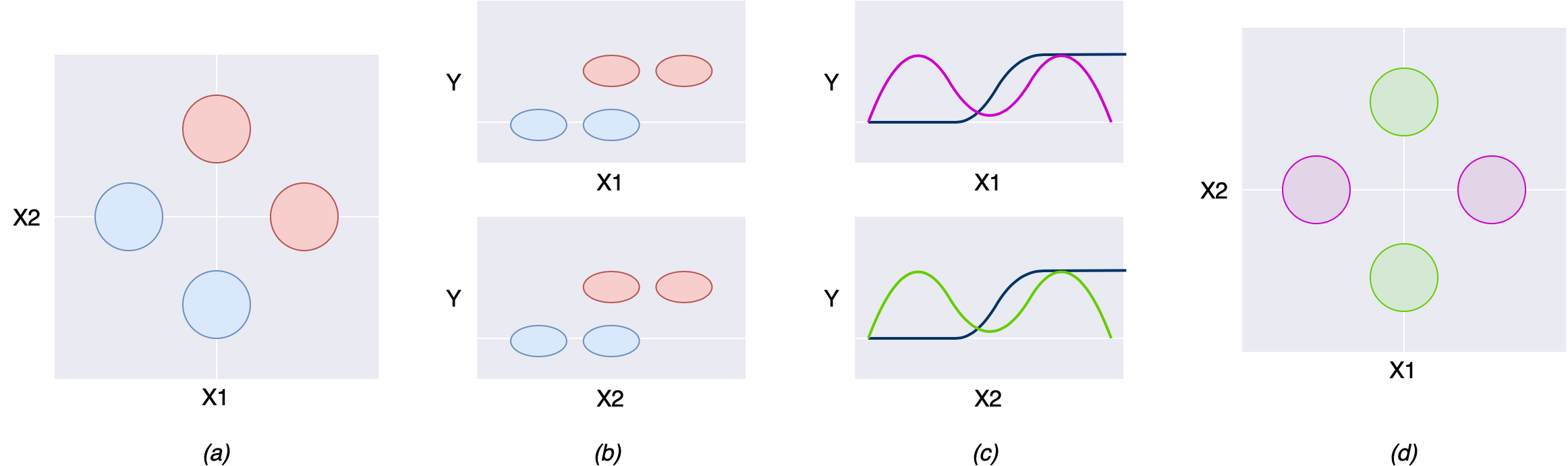}
  \caption{\textit{\textbf{Toy example (a)}} \textit{(a)} Input data with four clusters with 1 labels in red and 0 labels in blue ; \textit{(b)} restricted input data on the two dimensions $X_1$ and $X_2$ ; \textit{(c)} expected learned expert classification functions in black and selection functions in colours with high values where the clusters are separable ; \textit{(d)} resulting attribution with in purple an univariate expert based on $X_1$ and in green an expert based on $X_2$.}
  \label{toy_example_a}
\end{figure}

\begin{figure}[h]
\minipage{0.475\textwidth}
\centering
  \includegraphics[width=0.65\linewidth]{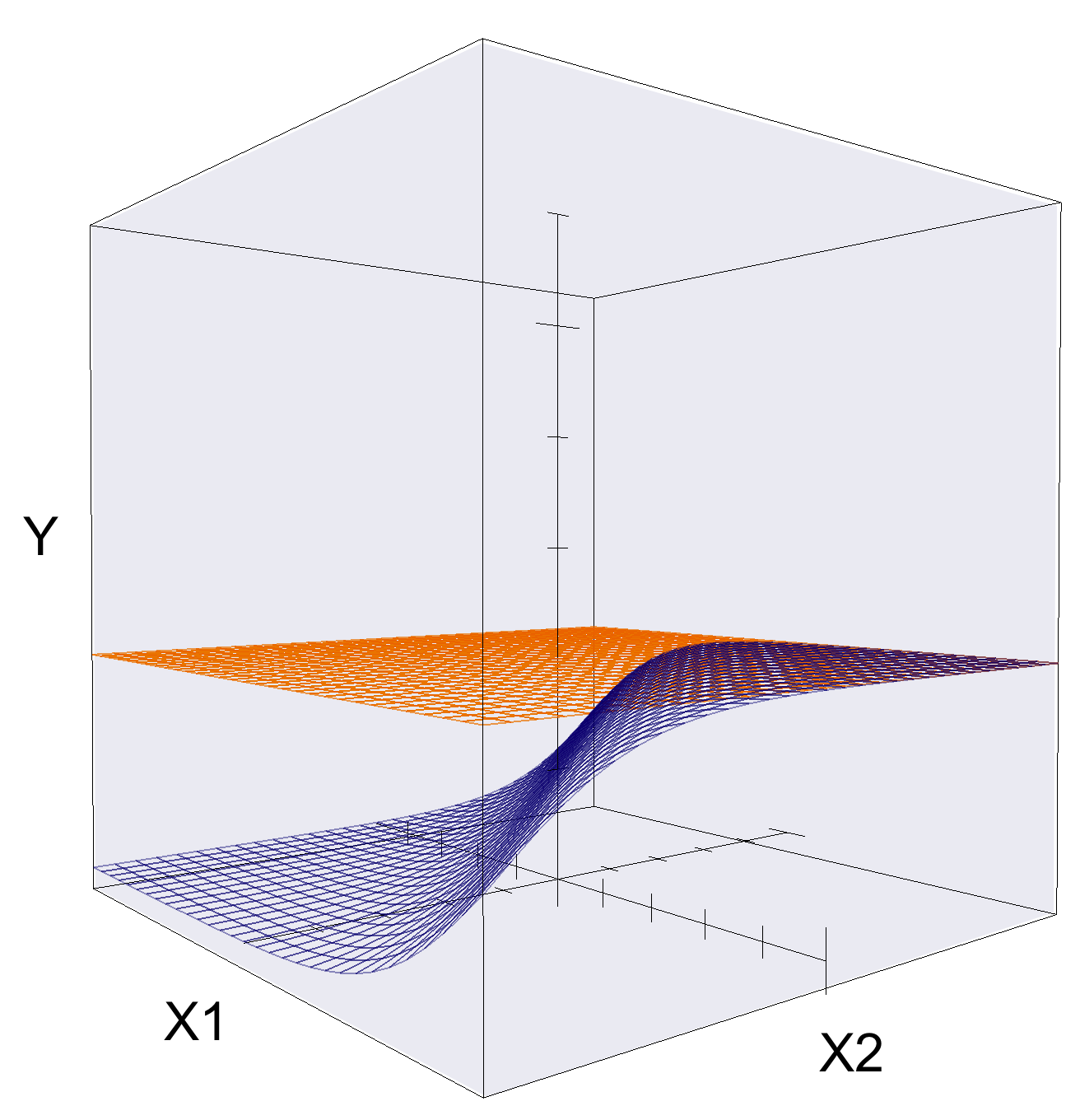}
  \caption{Bivariate solution for toy example (a), with expected classification function in black and selection function in orange. }\label{toy_example_a2}
\endminipage\hfill
\minipage{0.475\textwidth}
\centering
  \includegraphics[width=\linewidth]{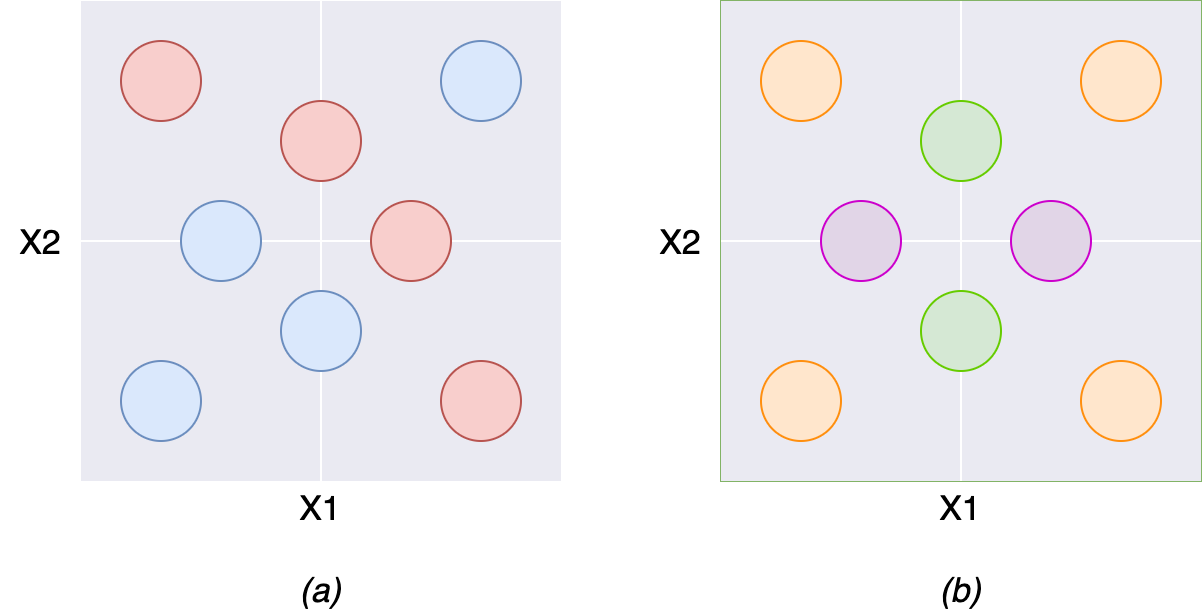}
  \caption{\textit{\textbf{Toy example (b)}} \textit{(a)} Input data that add four clusters to the first toy problem ; \textit{(b)} expected resulting attribution where the outer clusters are attributed to a new third expert based on $\{X_1, X_2 \}$  in orange.}\label{toy_example_b}
\endminipage
\end{figure}

To fix ideas, a bidimensional toy example with four input clusters is given in figure~\ref{toy_example_a}. We have illustrated a solution with two univariate experts, \textit{i.e.} $\mathcal{S} = \{S_1, S_2\} = \{\{X_1\}, \{X_2\}\}$: each predicts two separable clusters and their respective selection function have low values where the remaining clusters are mixed. A bivariate expert ($S_3 = \{X_1, X_2\}$) can also solve the task (\textit{fig.}
~\ref{toy_example_a2}), but in order to have the sparsest $M$, \textbf{we would favour the first solution} and have a zero-selection on the bivariate expert. In figure \ref{toy_example_b}, we add four outer clusters to example \textit{(a)} that are not separable when projected on $X_1$ nor $X_2$. In this case, the new clusters have to be attributed to the bivariate expert. A single bivariate expert could solve the whole task alone, but in order to get the sparsest mask $M$, denoting the minimal required input features to make a correct prediction in a specific part of the input space, using univariate experts is sufficient for the central clusters.

\subsubsection{Boosting}

In the general case, $\mathcal{S}$ is composed of many potentially overlapping subsets. We can represent subsets and experts with a directed acyclic graph (DAG) defined as the Hasse diagram of the subsets partially ordered by inclusions~\cite{bach2009high}: simple experts with few variables are children of parent experts of growing input support. The same way we select univariate experts over the bivariate expert in the toy example, the sparsity constraint means that the \textbf{selection of a child subset should induce the deselection of parent subset it is included in}. We ensure this property by design in a boosting-like manner: for each sample, we try to select an expert restricted to the smallest set of features, then, if it is not selected, we move to a parent expert. 

To allow training with a gradient-descent method, we consider a stochastic relaxation of the selection of the expert. We introduce the parametric functions $g^i_\theta : S_i \rightarrow [0, 1]$. We then define the selection functions recursively:
\begin{enumerate}
    \item \textbf{Atomic subsets:} For all $S_i \in \mathcal{S}$ such that $\forall S_j \in \mathcal{S} \setminus S_i, S_j \not\subset S_i$, then $\alpha^i_\theta(x) = g^i_\theta(x \circ m_i)$ ;
    \item \textbf{Mixed subsets:} For the remaining subsets, we introduce the notation $\omega(i) = \{ j | S_j \subsetneq S_i \}$ for the set of strictly included subsets and $\alpha^i_\theta(x) = g^i_\theta(x \circ m_i) \prod_{j \in \omega(i)} (1 - g^j_\theta(x \circ m_j))$
\end{enumerate}

By induction, the functions $\alpha^i$ we design satisfy the following properties:

\begin{enumerate}
    \item \textbf{Probability:} $\alpha^i \in [0, 1]$
    \item \textbf{Input restriction dependence:} $\alpha^i$ is entirely conditioned on $S_i$, and is thus blind to eventual parent subsets
    \item \textbf{Deselection induced by children:}  $\alpha^i \leq \prod_{j \in \omega(i)} (1 - g^j_\theta)$
\end{enumerate}


\subsubsection{Parametrisation using a neural network}

There are many possible choices for $f^i_\theta$ and $g^i_\theta$. In the rest of the paper, we study the use of deep neural networks, that have good generalisation capabilities, and their specific adaptation to our interpretation framework. Let us denote $F_\theta^i$ a deep neural network function restricted on $S_i$.



For the binary problems, we propose to use a \textit{tanh} function on the output layer. Then, we use the joint definition:
\begin{center}
\begin{minipage}{0.475\linewidth}
    \begin{equation}
        \label{eq:fi}
        f^i_\theta(x) = (F_\theta^i(x) + 1)/2
    \end{equation}
\end{minipage}
\hfill
\begin{minipage}{0.475\linewidth}
    \begin{equation}
       g^i_\theta(x) = |F_\theta^i(x)|
    \end{equation}
\end{minipage}
\end{center}

The neural network output simultaneously makes a prediction for $Y$ and its absolute value indicates a confidence value for selection as an expert. We experimentally found that for inference, using $g^i_\theta(x) = F_\theta^i(x)^2$ or $g^i_\theta(x) = \frac{2|F_\theta^i(x)|^p}{1 + |F_\theta^i(x)|^{p-1}}$ also worked well to dampen noisy values around 0 then smoothly increasing selection importance for stronger predictions.

\subsubsection{Training}
\label{sec:training}

We could sequentially maximise the likelihood for each expert with subsets of increasing cardinality, or even group independent subsets for fewer training phases, as in \cite{lou2013accurate} where all univariate functions are trained in parallel before bivariate functions. This approach can be time-consuming, especially with neural networks as experts.

Instead, \textbf{we train all experts in parallel} using EM. However, with neural networks, we do not have a tractable solution for $\arg \max_\theta \mathcal{Q}(\theta, \theta_{\textrm{old}})$. This issue is addressed by \ac{GEM} by substituting the maximisation with an incremental update of $\mathcal{Q}$.
The training of our models follows two alternating steps:
\begin{enumerate}
    \item \textbf{E-step} Evaluate $p_{\theta^\textrm{old}}(m_i|x) = \alpha^i_{\theta^\textrm{old}}(x \circ m_i) = |F^i_{\theta^\textrm{old}}(x \circ m_i)| \prod_{j \in \omega(i)}(1 - |F^j_{\theta^\textrm{old}}(x \circ m_j)|) $, which weights the sample $x$ differently for each expert with a deselection for parents ;
    \item \textbf{Generalised M-step} Perform a gradient-step update: $\theta_\textrm{new} = \theta_{\textrm{old}} + \eta \frac{\partial \mathcal{Q}}{\partial \theta}(\theta, \theta_{\textrm{old}})$, with $\eta$ the learning rate.
\end{enumerate}
Following the derivation of $\mathcal{Q}$ in equation~\eqref{eq:binary_derivation}, and using equation \eqref{eq:fi}, we have:
\begin{small} 
\begin{equation}
    \frac{\partial \mathcal{Q}}{\partial \theta}(\theta, \theta_{\textrm{old}}) =
    -\mathbb{E}_{x \sim X, y \sim Y}\left[ \sum_{m \in \mathcal{M}} \alpha^m_{\theta_\textrm{old}}(x \circ m) \frac{\partial \textrm{BCE}}{\partial \theta} \left( y, \frac{F^m_\theta(x \circ m) + 1}{2} \right) \right]
    \label{eq:loss}
\end{equation}
\end{small}
The M-step can be easily implemented in modern deep learning libraries to propagate the updates through the layers of the experts $F_\theta^m$. In the next section, we detail their architecture.

\subsection{Making neural networks interpretable}
\label{sec:nn}

So far, we have only considered the $H$ experts as being distinct entities. We show that assuming that the collections $(f^i_\theta)$ and $(g^i_\theta)$ are based on a neural network functions $(F^i_\theta)$,  everything can be grouped into a single deep neural network. We formulate such neural network by induction, assuming the conventional multi-layered architecture that alternates linear layers and activation functions. We then extend our method to a broader class of deep models.


\subsubsection{One-layer neural network}

We assume that functions $(F^i_\theta)$ have a single linear layer with an activation function $\sigma$:
\begin{align}
    F^i_\theta(x) = \sigma(W^i_0(x \circ m_i)) = \sigma(\tilde W^i_0 x )
\end{align}
We suppose we have added a scalar $1$ to the input $x$ to account for bias when multiplying by matrix $(W^i_0)$ to simplify notations. $\tilde W^i_0$ corresponds to matrix $W^i_0$ with null columns at the indices where $m_i$ is null, \textit{i.e.} to $W^i_0 \circ m_i$ with the Hadamard product applied row by row. Then, we can stack the matrices and define $F_\theta(x)$:
\begin{small}
\begin{align}
    F_\theta(x)
    = \begin{bmatrix} F^1_\theta (x) ... & F^H_\theta (x) \end{bmatrix}^T
    = \sigma \left( \begin{bmatrix} \tilde W_0^1 ... & \tilde W_0^H \end{bmatrix}^T x \right)
     = \sigma ( \tilde W_0 x )
\end{align}
\end{small}
We identify $F_\theta$ as an overarching single-layer neural network with activation $\sigma$ and matrix parameter $\tilde W_0$. The latter matrix is typically sparse because of the masks $(m_i)$ successively applied on each row and can be efficiently implemented using a weight constraint in standard deep learning libraries. We have shown the base case: using masks, \textbf{we can create one-layer networks for which the output dependencies to interpretation subsets can be traced}. We must now prove by induction that it can be extended to several stacked layers.

\subsubsection{Multi-layer neural network}

When functions $(F^i_\theta)$ are multi-layered neural networks with $K$ layers, activation functions $\sigma_k$, matrices parameters $W_k^i$ and hidden layer output $h^i_k$ on layer $k$, where $h^i_K \triangleq F^i_\theta$, we have by definition:
\begin{align}
    h^i_0(x) = & \sigma_0(W_0^i (x \circ m_i)) = \sigma_0(\tilde W^i_0 x ) \\
    h^i_{k+1}(x) = &  \sigma_k(W_k^i h^i_k(x)) \quad \forall k \in [ 0.. K-1 ]
\end{align}

Our goal is to define overarching hidden layers $h_{k}$ that are conditioned on the corresponding restriction $S_i$ for $k > 0$:
\begin{align}
    h_{k}
    = \begin{bmatrix} x \circ m_1 \rightarrow h^1_{k}(x) & \dots &
    x \circ m_H \rightarrow h^H_{k}(x) \end{bmatrix}^T
\end{align}
Let us assume we have already built interpretable hidden layers up to layer $k$. As in the case of the first layer, we would like to define a masked matrix $W_k$ such that $h_{k+1}(x) = \sigma_k(W_k h_k(x))$, while preserving the correct input dependencies.

A first approach is to define $W_k$ as a diagonal by blocks with submatrices $W_k^i$:
\begin{small} 
\begin{align}
    h_{k+1}(x)
    = \begin{bmatrix} \sigma_k (W_k^1 h^1_k(x)) \\ \vdots \\ \sigma_k (W_k^H h^H_k(x)) \end{bmatrix}
    = \sigma_k \left(
        \begin{bmatrix} W^1_k & 0 & ... & 0 \\
            \vdots &  & & \vdots \\ 0 & ... & 0 & W^H_k \end{bmatrix}
        \begin{bmatrix} h^1_k(x) \\ \vdots \\ h^H_k(x) \end{bmatrix} \right)
    =  \sigma(W_k h_k(x))
\end{align}
\end{small}
We would again use a masked matrix with sparse parameters and it would be equivalent to having $H$ neural networks trained in parallel but yet remaining independent from one another because the hidden features of each expert is only used in the computation of its corresponding upper hidden layer features.

However, we do not change the desired property of dependencies by also allowing to \textbf{compute the hidden layer $h_{k+1}^i$ using $h_k^j$ for all $j \in \omega(i)$}. We thus rather define $W_k$ with non-null blocks $W_k^{i,j}$ everywhere $S_j \subset S_i$:
\begin{align}
    W_k
    = \begin{bmatrix}
        \textbf{1}_{S_1 \subset S_1} W^{1,1}_k & \dots & \textbf{1}_{S_j \subset S_1} W^{1,j}_k & \dots\\
        \vdots & & \vdots & \\
        \textbf{1}_{S_1 \subset S_H} W^{H,1}_k & \dots & \textbf{1}_{S_j \subset S_H} W^{H,j}_k & \dots\\ \end{bmatrix}
\end{align}
An example of such added links is given in the overview figure \ref{overview}.


The last step to be able to train this network using a gradient-descent-based algorithm is to \textbf{prevent back-propagation} before matrix blocks $W^{i,j}_k$ for $i \neq j$. Otherwise, the dependency is not preserved since child classifiers indirectly depend from their parents during training. This can be easily implemented in Tensorflow using a copy function as \texttt{stop\_gradient}. We have then recursively defined an \textbf{interpretable multi-layered neural network}.

\subsubsection{Extension}

In previous sections, we have formulated a simple way to create \textbf{interpretable linear layers} using masked matrices. Then, applying additional activations or element-wise operations (\textit{e.g.} skip connection, normalisation, ...) does not change the dependency of each sub-network $F^i_\theta$ on its restriction $S_i$. We can also apply functions along a time dimension to \textbf{extend our method to sequences of inputs}: as long as we process together hidden features computed using the same expert ($i$) or its child experts ($j \in \omega(i)$), we do not change the restricted input dependency. 

An interesting case is the Transformer model~\cite{vaswani2017attention} or its variants~\cite{devlin2019bert}, that have been recently popularised across many fields, often achieving state-of-the-art results. Those models leverage multi-head attention modules taking as input a query ($q$), key ($k$) and value ($v$):
\begin{align}
    \textrm{head}_h(q,k,v) = & \quad \textrm{DotProductAtt}(qW_h^q,kW_h^k,vW_h^v) \\
    \textrm{MultiHead}(q,k,v) = & \quad \textrm{Concat}_{h=0}^P(\textrm{head}_h)W^o
\end{align}
Defining $P$ as a multiple of $H$, we partition the heads into several groups that act as experts. With the same procedure as before, we constraint parameters $W^q, W^k, W^v, W^o$ to be masked matrices to obtain an \textbf{interpretable multi-head attention module}. The remaining building blocks of the Transformer are element-wise functions or can be made interpretable, allowing to formulate an \textbf{interpretable Transformer} model.

Other deep architectures can be made interpretable using the same principle. We have derived an \textbf{interpretable gated recurrent units network} during our experiments. A bit of thinking can be required to correctly link each sub-function correctly in complex cases, for instance when using multiple heterogeneous inputs. Several implementation examples can be found on this paper code repository for more details~\footnote{\label{code_repo} Our code repository: \url{https://github.com/deezer/interpretable_nn_attribution}}.

\section{Experimental Results}

\label{sec:4}

We evaluate our method with the following research questions:
\begin{itemize}
    \item \textbf{RQ1} Do our deep interpretable models perform as well as their non-interpretable counterpart? (\textit{completeness})
    \item \textbf{RQ2} Are the provided interpretation relevant? (\textit{interpretability})
\end{itemize}

Interpretation is task-dependent, we thus study several implicit signals prediction tasks to see how our method fares in various settings: on toy example \textit{(b)} for which target attributions are available, on a collaborative filtering task using the \textit{MovieLens} dataset, and on the \textit{sequential skip prediction} task using user log data from Spotify and Deezer.

\subsection{Synthetic data}
\label{sec:exp_synthetic}

\subsubsection{Setting}
We simulate a mixture of eight Gaussian distributions according to \textbf{toy task \textit{(b)}} we have introduced in section \ref{sec:selection_dep}. With $X = (X_1, X_2)$ and $Y \in \{-1, 1\}$, we define three interpretation subsets $S_1, S_2, S_3 = \{X_1\}, \{X_2\}, \{X_1, X_2\}$, \textit{i.e.} two univariate experts, $f^1$ and $f^2$, and one residual bivariate expert, $f^3$. We instantiate an interpretable two-layers feed-forward network with $3 \times 16$ neurons on each hidden layer and \textit{ReLU} activations and train it for a few minutes until convergence using \textit{Adam}~\cite{kingma2014adam} with default parameters\textsuperscript{\ref{code_repo}}.

\subsubsection{Results} 
\begin{figure*}[h]
  \centering
  \includegraphics[width=\linewidth]{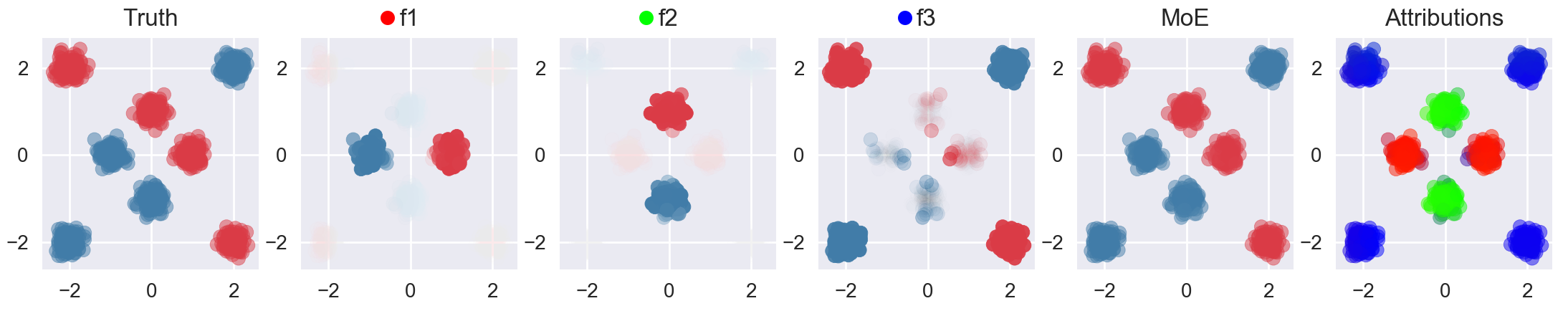}
  \caption{\textit{\textbf{Results on toy task (b)}} A sample input for toy task \textit{(b)} is highlighted in the leftmost figure. Predicted labels are plotted with a red colour for $1$ and blue for $-1$. Selection values were passed in the alpha channel to highlight the attribution behaviour. In the last plot, the selection values were passed to each colour channel: red $\sim S_1$, green $\sim S_2$, blue $\sim S_3$. }
  \label{res_toy_example}
\end{figure*}

Except for a few misclassified edges points, this task is simple enough to be almost perfectly solved by the network, as shown in figure~\ref{res_toy_example} (\textbf{RQ1}). We see that the expected attribution (\textit{fig.} \ref{toy_example_b}) is obtained, the central clusters are attributed to each univariate expert instead of using the expert with all input features (\textbf{RQ2}).

\subsection{Collaborative filtering on implicit signals}
\label{sec:exp_ml}

\subsubsection{Dataset}
We evaluate our method on a more realistic \textit{collaborative filtering} task (CF). We reproduce the setup of \textit{NCF}~\cite{he2017neural} and compare their method with an equivalent interpretable network. Specifically, we use the \textit{MovieLens 1M} dataset~\footnote{\url{https://grouplens.org/datasets/movielens/1m/}}, containing one million movie ratings from around six thousands users and four thousands items. All ratings are binarised as implicit feedbacks to mark a positive user-item interaction, while non-interacted items are considered negative feedbacks. The performance is evaluated with the \textit{leave-one-out} procedure, and judged by \textit{Hit rate} (HR) and \textit{Normalised Discounted Cumulative Gain} (NDCG). Validation is done by isolating a random training item for each user. More details can be found in the original paper
~\cite{he2017neural} and \cite{dacrema2019we}.

\subsubsection{Interpretation setting}
In CF, a user $u$ (\textit{resp.} item $i$) is typically embedded into a latent vector $p_u$ (\textit{resp.} $q_i$), and the observed interactions are estimated via a similarity function. In \textit{NCF}, the authors propose to replace the traditional inner product by a neural network to compute similarities. Because of the projection on a latent space, CF is more difficult to interpret than a \textit{content-based} method that would only leverage the provided descriptive features for users - \texttt{age range}, \texttt{gender}, \texttt{occupation} ($c_u$) - and movies - \texttt{year}, \texttt{genres} ($c_i$). A model merely treating user and item using generic ranges (\textit{i.e.} clusters) instead of personalised embeddings is however too coarse and underperforms.

Here, our method can be used to \textbf{mix content-based and CF experts} to discriminate interactions that can be predicted based on content, from the one that need an additional CF treatment to model users and items particularities. This way, we can trace if an item is recommended because of its similarity among a generic item range (\textit{eg.} similar to \textit{horror-movies from the 90's}), a user range (\textit{eg.} also liked by \textit{male viewers in their twenties}), the combination of both, or beyond using CF. To this end, we define four experts with $\mathcal{S} = S_1,\dots S_4 = \{c_u, c_i\}, \{c_u, p_u, c_i\}, \{c_u, c_i, q_i\}, \{c_u, p_u, c_i, q_i \}$.

We use the multi-layered network version of \textit{NCF} named \textit{MLP} in \cite{he2017neural}, and parametrise $p_u, q_i \in \mathbb{R}^{64}$, $c_u, c_i \in \mathbb{R}^{16}$, and four hidden layer of sizes $[512, 256, 128, 64]$. The interpretable counterpart we dub \textit{Intrp-MLP}, is build with the same architecture but with masked weights\textsuperscript{\ref{code_repo}}, which is equivalent to having four experts with hidden layer sizes $[128, 64, 32, 16]$.
\subsubsection{Results}

\begin{figure}[h]
\begin{minipage}{0.475\textwidth}
\centering
\begin{tabular}{l|cc|c}
\toprule
Model & HR@10 & NDCG@10 & \#Params \\
\midrule
\textit{content-based} & 0.386 & 0.218 & 190K \\
\textit{NCF}-MLP & 0.715 & 0.438 & 890K \\
\midrule
\textit{Intrp}-MLP & 0.678 & 0.406 & 782K \\
\bottomrule
\end{tabular}
\captionof{table}{\textit{\textbf{Movielens-1M results}} As in \cite{he2017neural}, the metrics are computed by sampling 100 random negative items and ranking the left-out test element among them for each user. We include the \textit{content-based} model to highlight its underperformance.}
\label{table:movielens}
\end{minipage}
\hfill
\begin{minipage}{0.475\textwidth}
    \centering
    \includegraphics[width=\linewidth]{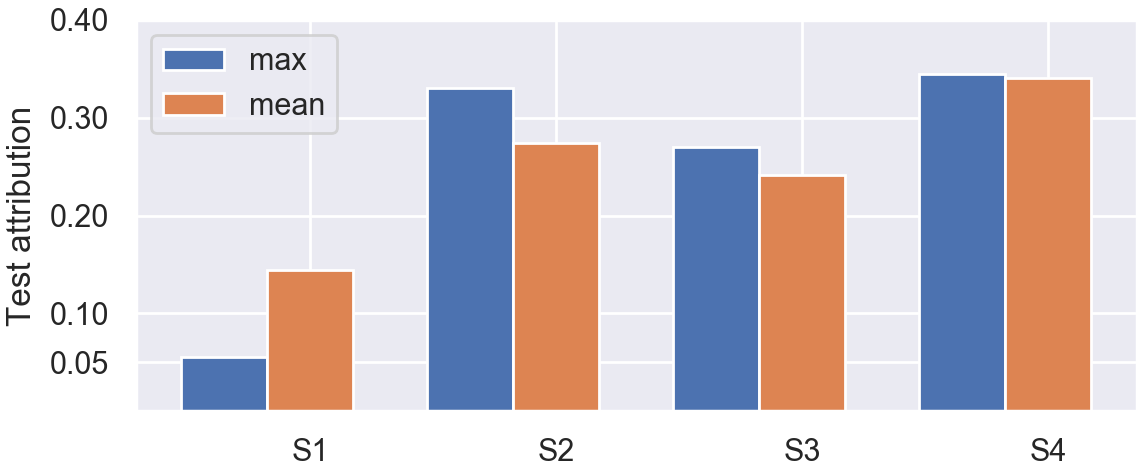}
    \captionof{figure}{\textit{\textbf{Movielens attribution}} Histogram of maximum-attributed experts (blue) and averaged attribution values (orange) over the test set items.}
    \label{fig:ml_attribution}
\end{minipage}
\end{figure}

As presented in table~\ref{table:movielens}, our interpretable version of \textit{NCF-MLP} achieves close performances to the control non-interpretable model, with a 4\% difference in \textit{HR} (\textbf{RQ1}). This latter control model has a better \textit{HR} and \textit{NDCG} than reported in \cite{he2017neural}, which can be explained by the addition of contextual features $c_u, c_i$ and bigger hidden layers.

Contrary to section \ref{sec:exp_synthetic}, we do not have access to ground-truth attributions to check our model interpretability. As a simple proxy, we can check the attribution distribution on the test set for \textbf{RQ2} (\textit{fig.} \ref{fig:ml_attribution}). A first sanity check is that attributions do not collapse to an unique expert and have a relative diversity. We also see that the pure content-based expert ($S_1$) is hardly selected, which is coherent with the underperformance of content-based models on this task.

Overall, we must underline that \textbf{66\%} of the items are predicted using the three first experts, for which \textbf{an interpretation can be provided} as either or both item and user will be described by generic features instead of a CF embedding: \textit{e.g.} the selection of $S_2$ (\textit{resp.} $S_3$) indicates a similarity to the item cluster (\textit{resp.} user), as movies from a specific year and genre. The residual 34\% are left for the CF expert ($S_4$) when further personalisation is needed.

\subsection{Sequential skip prediction}

\subsubsection{Dataset}
We study the task of \textit{skip prediction} with two music sessions datasets. First, the \textit{Music Streaming Sessions Dataset}~\cite{brost2019music}. This public dataset contains anonymised listening logs of users from the \textbf{Spotify} streaming service over an 8 week period. Listening logs are sequenced together for each user, forming roughly $150$ million listening sessions of length ranging from size 10 to 20. Sessions including unpopular tracks were excluded, limiting the overall track set to approximately 3.7 millions tracks.


For the \textit{sequential skip prediction} task\textsuperscript{\ref{wsdm}}, a session $(i)$ of length $2l^{(i)}$ is cut in half, with the first half (referred to as $A^{(i)} = A_1^{(i)} \dots  A_{l^{(i)}}^{(i)}$ $\in \mathbb{R}^{l^{(i)} \times f}$) containing sessions logs, user interactions logs and track metadata, while the second half ($B^{(i)} \in \mathbb{R}^{l^{(i)} \times g}$) contains only the track metadata. The goal is to predict the boolean value labelled as \texttt{skip\_2} ($Y^{(i)} \in \{-1, 1\}^{l^{(i)}}$) among the missing interaction features of $B^{(i)}$. We omit the indices $(i)$ to simplify notations.

In addition to this dataset, we also use a private streaming sessions dataset provided by the music streaming service \textbf{Deezer}. This dataset contains $10$ million listening sessions of range 20 to 50 from a week of streaming logs. To have a similar setup to Spotify, we extract random session slices of size 20 at each epoch, enabling to have virtually more listening sessions. Without the anonymity constraint, this private dataset enables to use more features and to better evaluate interpretations with tangible data that can be streamed and manually checked. This dataset notably includes user metadata, including \textit{favourite and banned tracks}, \textit{recently listened tracks}, \textit{mean skip rate}, \textit{user embedding}, \dots

The difficulty of the challenge lies in the multitude of origins a skip can have. For instance, we can check that users counterintuitively skip their favourite tracks with the almost same ratio as other tracks. Skips do not just signal a music a user does not like, they also happen when a track is liked by a user but streamed at the wrong time, or when a user quickly browses the catalog, looking for a specific song, or just fresh content, or with connection errors or misclicks. Conversely, \textit{non-skip} are also implicit, sometimes the user is simply not there to change an unwanted track.


\subsubsection{Metrics}
We compute the \textbf{accuracy} (Acc) of correctly predicted skip interaction in each half-session $B$. We also make use of the evaluation metrics introduced during the challenge, the \textbf{Mean Average Accuracy} (MAA), to allow for comparison. Average accuracy is defined by $AA = \sum_{j=1}^T \textrm{acc}(j)L(j)/T$, where $T$ is the number of track to be predicted in $B$, $\textrm{acc}(j)$ the accuracy at position $j$, and $L(j)$ a boolean indicating if the prediction at position $j$ was correct. This metric puts more weights on the first track of $B$ than the last ones.

It is argued that in the context of a session-based recommender system, this unbalance is due to the fact that it is more important to know if the next immediate track to be streamed will be skipped or not given preceding interactions and prevent a bad recommendation in the nearest future. However, as we will see, this latter argument can be flawed as first track prediction strongly depends on the blind continuation of interactions more than interesting underlying mechanisms that combine multiple features, hence not always providing much information to improve recommendation. We use the \textbf{accuracy on the first immediate track} (Acc@1) to highlight this effect. This question of relative relevance of a skip to a context needs to be addressed to allow retroactive improvements of a recommender system, which could be provided by the task we are trying to solve, \textit{skip interpretation}.


\subsubsection{Baseline}

As we suggested, skips strongly depend on the persistence of skip behaviours. This can be interpreted as users being active \textit{by blocks}: once a skip is performed, there is a higher chance that a user will also skip the next track while still on its app, and the other round, while not on the app, users may be more likely to tolerate an unwanted song. Because of this effect, it is relevant to use a \textit{persistence model} as baseline, returning the last known interaction in $A$ for all the elements of $B$. We additionally use a mean skip measure from $A$. We thus have two baseline predictors:

\begin{center}
\begin{minipage}{0.435\linewidth}
    \begin{equation}
        f_{\textrm{last}}(A, B) = A_l[\texttt{skip}]
    \end{equation}
\end{minipage}
\hfill
\begin{minipage}{0.555\linewidth}
    \begin{equation}
       f_{\textrm{mean}}(A, B) = \frac{1}{l} \sum_{j = 1}^{l} A_{j}[\texttt{skip}]
    \end{equation}
\end{minipage}
\end{center}


\subsubsection{Experimental setup}

We use a standard Transformer architecture~\cite{vaswani2017attention}, with 3 stacked identical self-attentive layer blocks for the encoder architecture with key-values input $A$, as in the original paper, and 3 cross-attentive layer blocks for the decoder with query input $B$. 

For interpretation, we define $8$ experts for the Spotify dataset (\textit{fig.} \ref{fig:spo_subsets}), and $10$ for Deezer (\textit{fig.} \ref{fig:deez_subsets}). It must be underlined that $A$ and $B$ do not have the same interpretation subsets on the first layer as some features are missing for $B$, the session to be predicted without logged interactions. The method however remains the same in this heterogeneous case to preserve dependencies: a link can be added between features based on $S_x$ and $S_y$ if and only if $S_y \subset S_x$.

We train all models using \textit{Adam} with default parameters and learning rate set to $10^{-4}$. The learning rate is automatically reduced on plateau up to $10^{-6}$. The two datasets are split in a 80-10-10\% fashion for training/validation/test.
Because of the huge number of sessions, the models reach convergence on the training loss in around roughly two epochs. We did not observe any overfitting effect, making it easy to control the optimisation. 

\subsubsection{Prediction performance (\textbf{RQ1})}

Results for both datasets are given in table \ref{table:res1} and \ref{table:res2}. Baselines models, though parameter-free, are performing strikingly well on both datasets, especially on the \textbf{first track} prediction where the continuation effect is the strongest. In both cases, our interpretable models have close performances to their non-interpretable architecture counterparts, though losing around a point of accuracy.


In the \textit{WSDM} Challenge\textsuperscript{\ref{wsdm}}, the evaluation was performed on a private and still inaccessible test set. However, the results of our baselines and control Transformer model seem to be coherent with the reported results, our control model would have been ranked to the fifth place of the challenge leaderboard.


\begin{table}[h]
\begin{center}
\begin{tabular}{l|ccc}
\toprule
Model & Acc (\%) & Acc@1 (\%) & MAA (\%) \\
\midrule
random & 50.0 & 50.0 & 33.1 \\
\textit{last} baseline & 63.0 & 74.2 & 54.3 \\
\textit{mean} baseline & 61.7 & 66.3 & 51.7 \\
\midrule
Transformer (128) & 72.2 $\pm$ 0.2 & 80.0 $\pm$ 0.2 & 62.8 $\pm$ 0.2 \\
\textit{WSDM leader-board}* & \textit{-} & \textit{81.2} & \textit{65.1} \\
\midrule
Interpretable Transformer (128) & 70.9 $\pm$ 0.2 & 78.8 $\pm$ 0.2 & 61.1 $\pm$ 0.2 \\
\bottomrule
\end{tabular}
\end{center}
\caption{\textit{\textbf{Spotify skip prediction test results}} Numbers in parenthesis indicate the size of the hidden layer.}
\label{table:res1}
\end{table}

\begin{table}[h]
\begin{center}
\begin{tabular}{l|ccc}
\toprule
Model & Acc (\%) & Acc@1 (\%) & MAA (\%) \\
\midrule
random & 50.0 & 50.0 & 32.3 \\
\textit{last} baseline & 69.0 & 77.9 & 60.8 \\
\textit{mean} baseline & 70.1 & 73.3 & 60.9 \\
\midrule
Transformer (256) & 78.9 & 83.4 & 70.2 \\
\midrule
Interpretable Transformer (128) & 77.7 & 82.4 & 68.8 \\
Interpretable Transformer (64) & 77.4 & 82.3 & 68.4 \\
\bottomrule
\end{tabular}
\end{center}
\caption{\textit{\textbf{Deezer skip prediction test results}} Numbers in parenthesis indicate the size of the hidden layer.}
\label{table:res2}
\end{table}

\subsubsection{Attribution distribution (\textbf{RQ2})}

\begin{figure}[h]
\minipage{0.465\textwidth}
\centering
  \includegraphics[width=0.9\linewidth]{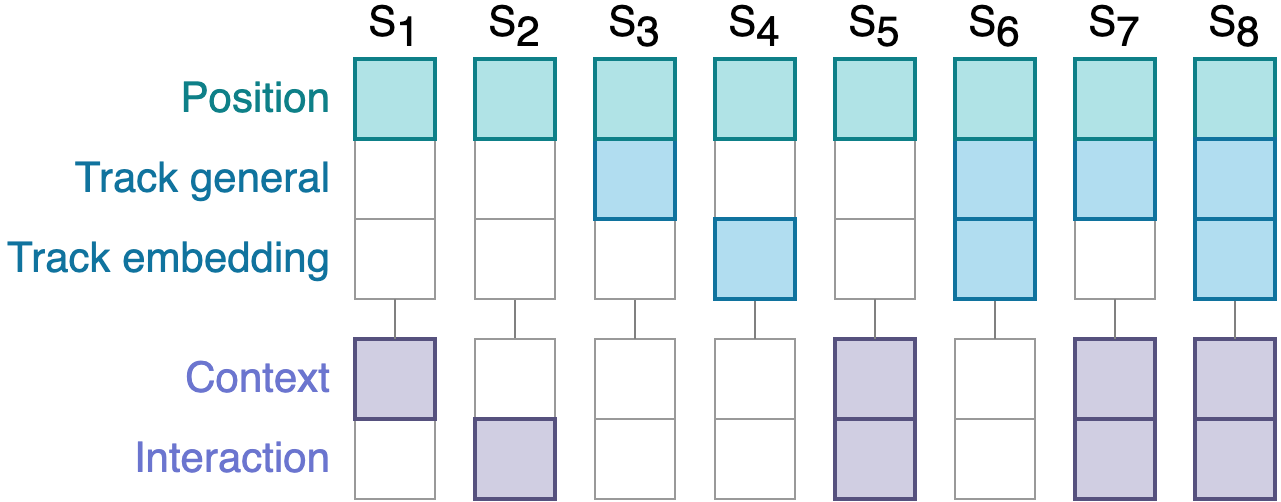}
  \caption{\textit{\textbf{Spotify interpretation subsets}} For each subset $S_i$, the feature groups we include (\textit{i.e.} $X_j$) are coloured. The groups of input features drawn in purple are not available in $B$.}
  \label{fig:spo_subsets}
\endminipage\hfill
\minipage{0.485\textwidth}
\centering
  \includegraphics[width=\linewidth]{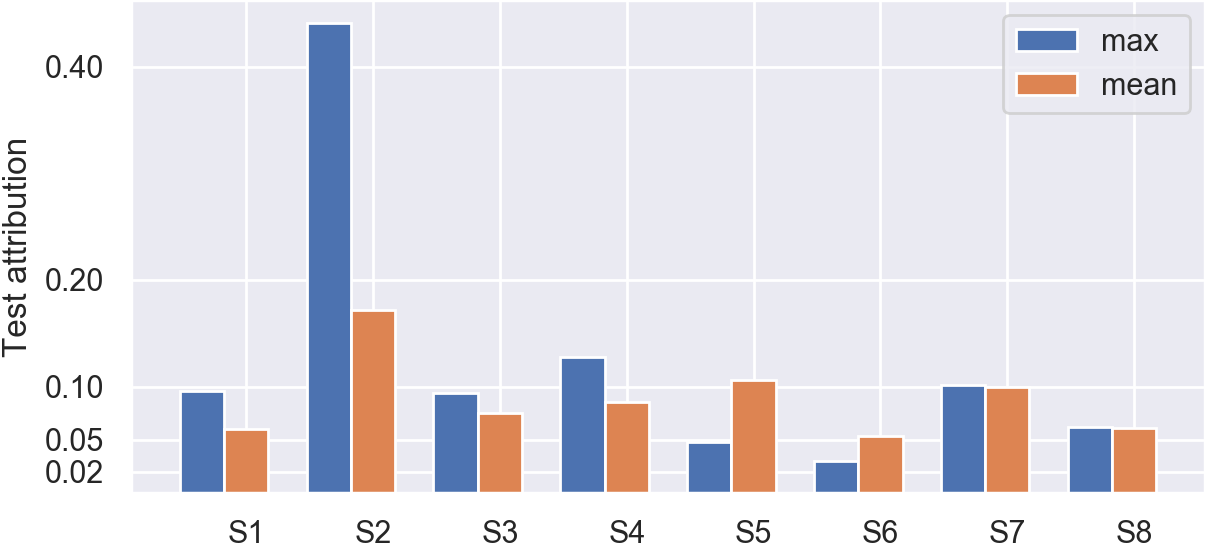}
  \caption{\textit{\textbf{Spotify attribution}} Mean attribution value and mean item-wise max skip attribution over $20'000$ test sessions.}
  \label{fig:res_spo}
\endminipage
\end{figure}

\begin{figure}[h]
\minipage{0.475\textwidth}
\centering
  \includegraphics[width=\linewidth]{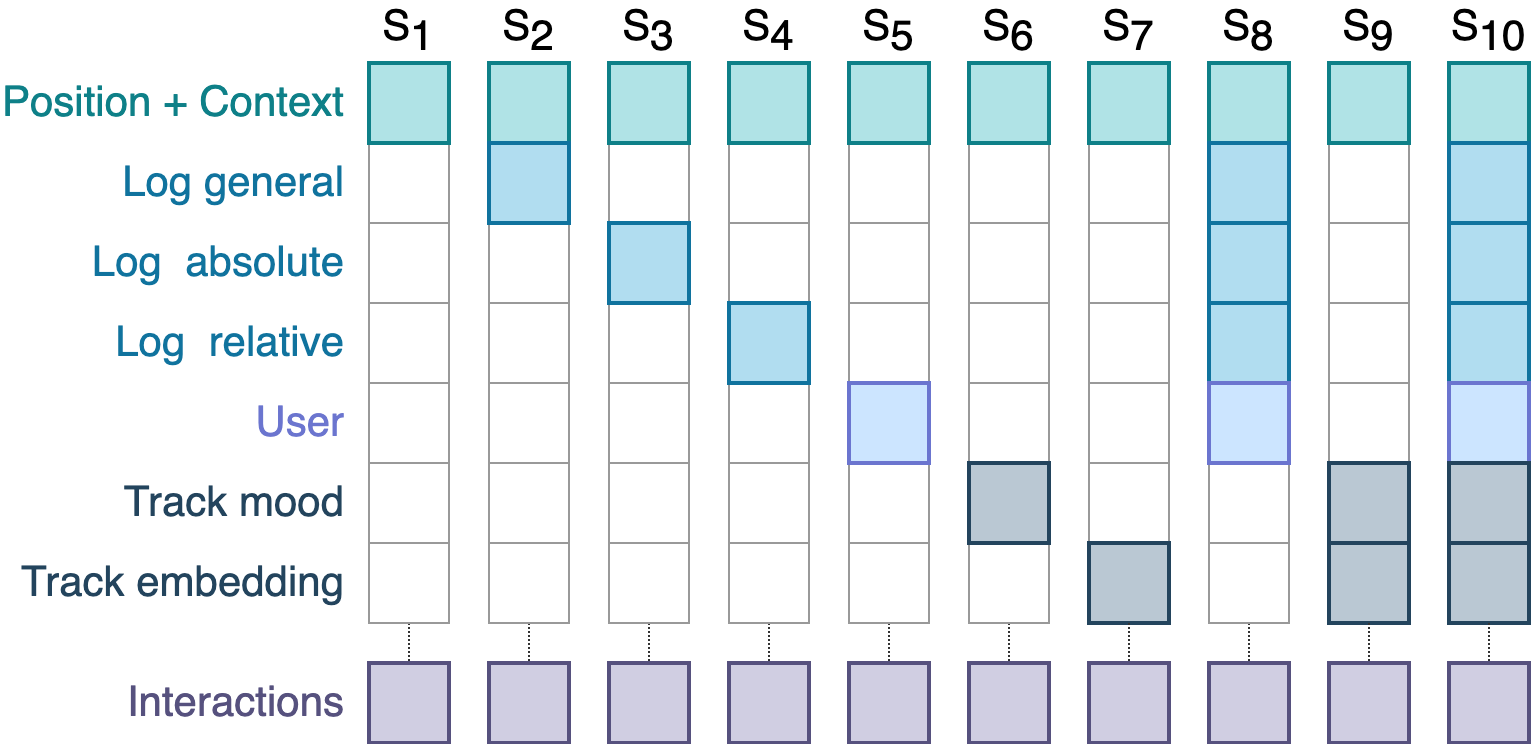}
  \caption{\textit{\textbf{Deezer interpretation subsets}} We mask all interactions input features in $B$ to preserve causality as in \cite{brost2019music}.}
  \label{fig:deez_subsets}
\endminipage\hfill
\minipage{0.475\textwidth}
\centering
  \includegraphics[width=\linewidth]{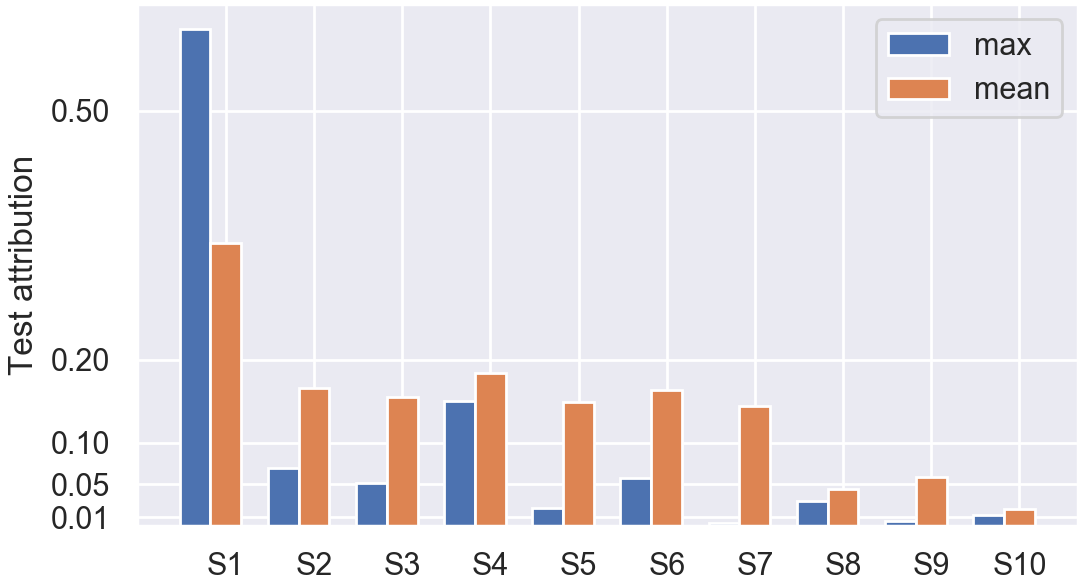}
  \caption{\textit{\textbf{Deezer attribution}} Mean attribution value and mean item-wise max skip attribution over $20'000$ test sessions.}
  \label{fig:res_deez}
\endminipage
\end{figure}

As in \ref{sec:exp_ml}, we inspect the attribution distribution on the test sessions of Spotify (\textit{fig.} \ref{fig:res_spo}) and Deezer (\textit{fig.} \ref{fig:res_deez}). In both cases, there is a strong unbalance toward the simplest expert containing the interaction data of $A$. This results allows to confirms our initial intuition that \textbf{most skips result from pure interaction patterns} and do not depend on other data. Those skips are not interesting for a recommender system as they do not tell much about user preferences. For the Spotify dataset, subsets $S_3$, $S_4$ and $S_6$ reveal that 25\% of skips can be predicted from the overall track metadata coherence, while being agnostic to the given skips in $A$, which hints at simple ways to filter tracks in a candidate recommended session continuation $B$.
For the Deezer dataset, the second most attributed expert ($S_4$) leverages a \textit{discounted skip rate} measure that indicates a recent user-track affinity. We can conclude from the attribution levels that this relative measure is a stronger indicator than a \textit{favourite track} signal ($S_2$) or \textit{overall popularity} ($S_3$) to predict skips.

\begin{figure*}[h]
  \centering
  \includegraphics[width=\linewidth]{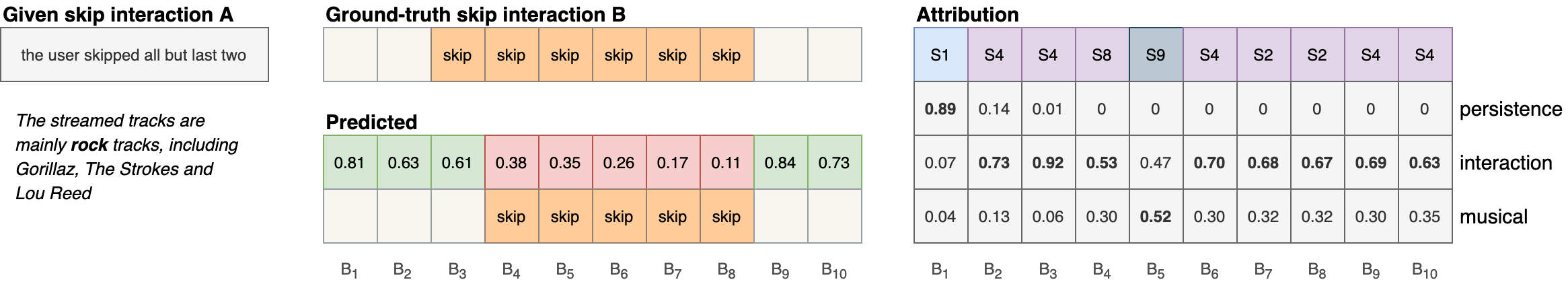}
  \caption{\textit{\textbf{Deezer skip prediction example}} We have aggregated the attribution values for concision: \textit{persistence} correspond to $S_1$, \textit{interaction} are the aggregation of $S_2$ to $S_5$ and $S_8$, \textit{musical} to $S_6$, $S_7$, $S_9$, the last expert with all features is hardly selected and hidden.}
  \label{fig:example_exp}
\end{figure*}

To illustrate the kind of interpretation we can get, an example of predicted session from Deezer is given in figure~\ref{fig:example_exp}. Quite typically, the $S_1$ expert that only contains the given skip in $A$ will have the strongest attribution for the first track of $B$, corresponding to the continuation of the last two non-skips in $A$, but will vanish rapidly for the next tracks in favour of more complex experts. In the middle of $B$, there is a Malaysian pop music, this rupture from the other rock songs can be observed in the sudden attribution to $S_9$, an expert based on musical data. Beyond simple cases of interaction continuation or favourite tracks, our method can handle this kind of multifactorial session and provide an insight on their nature.

\section{Conclusion}
\label{sec:5}

We introduce a novel attribution method that provides intrinsic interpretability by formulating a mixture of restricted experts, where simple experts are prioritised over more complex ones. We evaluate our method on synthetic problems, for which a ground-truth local attribution is available for comparison, and real-data tasks, aiming at predicting and interpreting binary implicit signals. Our experiments demonstrate that not only our interpretable networks achieve similar performances as their non-interpretable counterparts, but also help produce coherent interpretations that can be used to better understand implicit data, and may be leveraged by recommender systems.

As mentioned, our main future direction is the extension of our method to learnable interpretation subsets, which are currently fixed as hyper-parameters. Prior to this subject, a deeper discussion regarding the expected properties and geometry of attribution solutions will be needed in the unsupervised case for local attribution methods.



\bibliographystyle{ACM-Reference-Format}
\bibliography{main}


\begin{thebibliography}{51}


\ifx \showCODEN    \undefined \def \showCODEN     #1{\unskip}     \fi
\ifx \showDOI      \undefined \def \showDOI       #1{#1}\fi
\ifx \showISBNx    \undefined \def \showISBNx     #1{\unskip}     \fi
\ifx \showISBNxiii \undefined \def \showISBNxiii  #1{\unskip}     \fi
\ifx \showISSN     \undefined \def \showISSN      #1{\unskip}     \fi
\ifx \showLCCN     \undefined \def \showLCCN      #1{\unskip}     \fi
\ifx \shownote     \undefined \def \shownote      #1{#1}          \fi
\ifx \showarticletitle \undefined \def \showarticletitle #1{#1}   \fi
\ifx \showURL      \undefined \def \showURL       {\relax}        \fi
\providecommand\bibfield[2]{#2}
\providecommand\bibinfo[2]{#2}
\providecommand\natexlab[1]{#1}
\providecommand\showeprint[2][]{arXiv:#2}

\bibitem[\protect\citeauthoryear{Alvarez-Melis and Jaakkola}{Alvarez-Melis and
  Jaakkola}{2018}]%
        {alvarez2018robustness}
\bibfield{author}{\bibinfo{person}{David Alvarez-Melis} {and}
  \bibinfo{person}{Tommi~S Jaakkola}.} \bibinfo{year}{2018}\natexlab{}.
\newblock \showarticletitle{On the robustness of interpretability methods}.
\newblock \bibinfo{journal}{\emph{Workshop on Human Interpretability in Machine
  Learning, ICML}} (\bibinfo{year}{2018}), \bibinfo{pages}{66--71}.
\newblock


\bibitem[\protect\citeauthoryear{Bach}{Bach}{2009}]%
        {bach2009high}
\bibfield{author}{\bibinfo{person}{Francis Bach}.}
  \bibinfo{year}{2009}\natexlab{}.
\newblock \showarticletitle{High-dimensional non-linear variable selection
  through hierarchical kernel learning}.
\newblock \bibinfo{journal}{\emph{arXiv preprint arXiv:0909.0844}}
  (\bibinfo{year}{2009}).
\newblock


\bibitem[\protect\citeauthoryear{Bach, Binder, Montavon, Klauschen, M{\"u}ller,
  and Samek}{Bach et~al\mbox{.}}{2015}]%
        {bach2015pixel}
\bibfield{author}{\bibinfo{person}{Sebastian Bach}, \bibinfo{person}{Alexander
  Binder}, \bibinfo{person}{Gr{\'e}goire Montavon}, \bibinfo{person}{Frederick
  Klauschen}, \bibinfo{person}{Klaus-Robert M{\"u}ller}, {and}
  \bibinfo{person}{Wojciech Samek}.} \bibinfo{year}{2015}\natexlab{}.
\newblock \showarticletitle{On pixel-wise explanations for non-linear
  classifier decisions by layer-wise relevance propagation}.
\newblock \bibinfo{journal}{\emph{PloS one}} \bibinfo{volume}{10},
  \bibinfo{number}{7} (\bibinfo{year}{2015}).
\newblock


\bibitem[\protect\citeauthoryear{Bishop}{Bishop}{2006}]%
        {bishop2006pattern}
\bibfield{author}{\bibinfo{person}{Christopher~M Bishop}.}
  \bibinfo{year}{2006}\natexlab{}.
\newblock \bibinfo{booktitle}{\emph{Pattern recognition and machine learning}}.
\newblock \bibinfo{publisher}{springer}.
\newblock


\bibitem[\protect\citeauthoryear{Brost, Mehrotra, and Jehan}{Brost
  et~al\mbox{.}}{2019}]%
        {brost2019music}
\bibfield{author}{\bibinfo{person}{Brian Brost}, \bibinfo{person}{Rishabh
  Mehrotra}, {and} \bibinfo{person}{Tristan Jehan}.}
  \bibinfo{year}{2019}\natexlab{}.
\newblock \showarticletitle{The music streaming sessions dataset}. In
  \bibinfo{booktitle}{\emph{The World Wide Web Conference}}.
  \bibinfo{pages}{2594--2600}.
\newblock


\bibitem[\protect\citeauthoryear{Chang, Lee, and Lee}{Chang
  et~al\mbox{.}}{2019}]%
        {chang2019sequential}
\bibfield{author}{\bibinfo{person}{Sungkyun Chang}, \bibinfo{person}{Seungjin
  Lee}, {and} \bibinfo{person}{Kyogu Lee}.} \bibinfo{year}{2019}\natexlab{}.
\newblock \showarticletitle{Sequential Skip Prediction with Few-shot in
  Streamed Music Contents}.
\newblock \bibinfo{journal}{\emph{arXiv preprint arXiv:1901.08203}}
  (\bibinfo{year}{2019}).
\newblock


\bibitem[\protect\citeauthoryear{Chen, Duan, Houthooft, Schulman, Sutskever,
  and Abbeel}{Chen et~al\mbox{.}}{2016}]%
        {chen2016infogan}
\bibfield{author}{\bibinfo{person}{Xi Chen}, \bibinfo{person}{Yan Duan},
  \bibinfo{person}{Rein Houthooft}, \bibinfo{person}{John Schulman},
  \bibinfo{person}{Ilya Sutskever}, {and} \bibinfo{person}{Pieter Abbeel}.}
  \bibinfo{year}{2016}\natexlab{}.
\newblock \showarticletitle{Infogan: Interpretable representation learning by
  information maximizing generative adversarial nets}. In
  \bibinfo{booktitle}{\emph{Advances in neural information processing
  systems}}. \bibinfo{pages}{2172--2180}.
\newblock


\bibitem[\protect\citeauthoryear{Dacrema, Cremonesi, and Jannach}{Dacrema
  et~al\mbox{.}}{2019}]%
        {dacrema2019we}
\bibfield{author}{\bibinfo{person}{Maurizio~Ferrari Dacrema},
  \bibinfo{person}{Paolo Cremonesi}, {and} \bibinfo{person}{Dietmar Jannach}.}
  \bibinfo{year}{2019}\natexlab{}.
\newblock \showarticletitle{Are we really making much progress? A worrying
  analysis of recent neural recommendation approaches}. In
  \bibinfo{booktitle}{\emph{Proceedings of the 13th ACM Conference on
  Recommender Systems}}. \bibinfo{pages}{101--109}.
\newblock


\bibitem[\protect\citeauthoryear{Devlin, Chang, Lee, and Toutanova}{Devlin
  et~al\mbox{.}}{2019}]%
        {devlin2019bert}
\bibfield{author}{\bibinfo{person}{Jacob Devlin}, \bibinfo{person}{Ming-Wei
  Chang}, \bibinfo{person}{Kenton Lee}, {and} \bibinfo{person}{Kristina
  Toutanova}.} \bibinfo{year}{2019}\natexlab{}.
\newblock \showarticletitle{BERT: Pre-training of Deep Bidirectional
  Transformers for Language Understanding}. In
  \bibinfo{booktitle}{\emph{Proceedings of the 2019 Conference of the North
  American Chapter of the Association for Computational Linguistics: Human
  Language Technologies, Volume 1 (Long and Short Papers)}}.
  \bibinfo{pages}{4171--4186}.
\newblock


\bibitem[\protect\citeauthoryear{Freund, Schapire, and Abe}{Freund
  et~al\mbox{.}}{1999}]%
        {freund1999short}
\bibfield{author}{\bibinfo{person}{Yoav Freund}, \bibinfo{person}{Robert
  Schapire}, {and} \bibinfo{person}{Naoki Abe}.}
  \bibinfo{year}{1999}\natexlab{}.
\newblock \showarticletitle{A short introduction to boosting}.
\newblock \bibinfo{journal}{\emph{Journal-Japanese Society For Artificial
  Intelligence}} \bibinfo{volume}{14}, \bibinfo{number}{771-780}
  (\bibinfo{year}{1999}), \bibinfo{pages}{1612}.
\newblock


\bibitem[\protect\citeauthoryear{Friedman, Hastie, and Tibshirani}{Friedman
  et~al\mbox{.}}{2001}]%
        {friedman2001elements}
\bibfield{author}{\bibinfo{person}{Jerome Friedman}, \bibinfo{person}{Trevor
  Hastie}, {and} \bibinfo{person}{Robert Tibshirani}.}
  \bibinfo{year}{2001}\natexlab{}.
\newblock \bibinfo{booktitle}{\emph{The elements of statistical learning}}.
  Vol.~\bibinfo{volume}{1}.
\newblock \bibinfo{publisher}{Springer series in statistics New York}.
\newblock


\bibitem[\protect\citeauthoryear{Hansen, Hansen, Alstrup, Simonsen, and
  Lioma}{Hansen et~al\mbox{.}}{2019}]%
        {hansen2019modelling}
\bibfield{author}{\bibinfo{person}{Christian Hansen}, \bibinfo{person}{Casper
  Hansen}, \bibinfo{person}{Stephen Alstrup}, \bibinfo{person}{Jakob~Grue
  Simonsen}, {and} \bibinfo{person}{Christina Lioma}.}
  \bibinfo{year}{2019}\natexlab{}.
\newblock \showarticletitle{Modelling Sequential Music Track Skips using a
  Multi-RNN Approach}. In \bibinfo{booktitle}{\emph{ACM International
  Conference on Web Search and Data Mining}}. Association for Computing
  Machinery.
\newblock


\bibitem[\protect\citeauthoryear{Hastie and Tibshirani}{Hastie and
  Tibshirani}{1990}]%
        {hastie1990generalized}
\bibfield{author}{\bibinfo{person}{Trevor~J Hastie} {and}
  \bibinfo{person}{Robert~J Tibshirani}.} \bibinfo{year}{1990}\natexlab{}.
\newblock \bibinfo{booktitle}{\emph{Generalized additive models}}.
  Vol.~\bibinfo{volume}{43}.
\newblock \bibinfo{publisher}{CRC press}.
\newblock


\bibitem[\protect\citeauthoryear{He, He, Song, Liu, Jiang, and Chua}{He
  et~al\mbox{.}}{2018}]%
        {he2018nais}
\bibfield{author}{\bibinfo{person}{Xiangnan He}, \bibinfo{person}{Zhankui He},
  \bibinfo{person}{Jingkuan Song}, \bibinfo{person}{Zhenguang Liu},
  \bibinfo{person}{Yu-Gang Jiang}, {and} \bibinfo{person}{Tat-Seng Chua}.}
  \bibinfo{year}{2018}\natexlab{}.
\newblock \showarticletitle{Nais: Neural attentive item similarity model for
  recommendation}.
\newblock \bibinfo{journal}{\emph{IEEE Transactions on Knowledge and Data
  Engineering}} \bibinfo{volume}{30}, \bibinfo{number}{12}
  (\bibinfo{year}{2018}), \bibinfo{pages}{2354--2366}.
\newblock


\bibitem[\protect\citeauthoryear{He, Liao, Zhang, Nie, Hu, and Chua}{He
  et~al\mbox{.}}{2017}]%
        {he2017neural}
\bibfield{author}{\bibinfo{person}{Xiangnan He}, \bibinfo{person}{Lizi Liao},
  \bibinfo{person}{Hanwang Zhang}, \bibinfo{person}{Liqiang Nie},
  \bibinfo{person}{Xia Hu}, {and} \bibinfo{person}{Tat-Seng Chua}.}
  \bibinfo{year}{2017}\natexlab{}.
\newblock \showarticletitle{Neural collaborative filtering}. In
  \bibinfo{booktitle}{\emph{Proceedings of the 26th international conference on
  world wide web}}. \bibinfo{pages}{173--182}.
\newblock


\bibitem[\protect\citeauthoryear{Hendricks, Akata, Rohrbach, Donahue, Schiele,
  and Darrell}{Hendricks et~al\mbox{.}}{2016}]%
        {hendricks2016generating}
\bibfield{author}{\bibinfo{person}{Lisa~Anne Hendricks},
  \bibinfo{person}{Zeynep Akata}, \bibinfo{person}{Marcus Rohrbach},
  \bibinfo{person}{Jeff Donahue}, \bibinfo{person}{Bernt Schiele}, {and}
  \bibinfo{person}{Trevor Darrell}.} \bibinfo{year}{2016}\natexlab{}.
\newblock \showarticletitle{Generating visual explanations}. In
  \bibinfo{booktitle}{\emph{European Conference on Computer Vision}}. Springer,
  \bibinfo{pages}{3--19}.
\newblock


\bibitem[\protect\citeauthoryear{Herlocker, Konstan, and Riedl}{Herlocker
  et~al\mbox{.}}{2000}]%
        {herlocker2000explaining}
\bibfield{author}{\bibinfo{person}{Jonathan~L Herlocker},
  \bibinfo{person}{Joseph~A Konstan}, {and} \bibinfo{person}{John Riedl}.}
  \bibinfo{year}{2000}\natexlab{}.
\newblock \showarticletitle{Explaining collaborative filtering
  recommendations}. In \bibinfo{booktitle}{\emph{Proceedings of the 2000 ACM
  conference on Computer supported cooperative work}}.
  \bibinfo{pages}{241--250}.
\newblock


\bibitem[\protect\citeauthoryear{Hidasi, Karatzoglou, Baltrunas, and
  Tikk}{Hidasi et~al\mbox{.}}{2016}]%
        {hidasi2015session}
\bibfield{author}{\bibinfo{person}{Bal{\'a}zs Hidasi},
  \bibinfo{person}{Alexandros Karatzoglou}, \bibinfo{person}{Linas Baltrunas},
  {and} \bibinfo{person}{D Tikk}.} \bibinfo{year}{2016}\natexlab{}.
\newblock \showarticletitle{Session-based recommendations with recurrent neural
  networks}. In \bibinfo{booktitle}{\emph{4th International Conference on
  Learning Representations, ICLR 2016}}.
\newblock


\bibitem[\protect\citeauthoryear{Hinton, Krizhevsky, and Wang}{Hinton
  et~al\mbox{.}}{2011}]%
        {hinton2011transforming}
\bibfield{author}{\bibinfo{person}{Geoffrey~E Hinton}, \bibinfo{person}{Alex
  Krizhevsky}, {and} \bibinfo{person}{Sida~D Wang}.}
  \bibinfo{year}{2011}\natexlab{}.
\newblock \showarticletitle{Transforming auto-encoders}. In
  \bibinfo{booktitle}{\emph{International conference on artificial neural
  networks}}. Springer, \bibinfo{pages}{44--51}.
\newblock


\bibitem[\protect\citeauthoryear{Hu, Koren, and Volinsky}{Hu
  et~al\mbox{.}}{2008}]%
        {hu2008collaborative}
\bibfield{author}{\bibinfo{person}{Yifan Hu}, \bibinfo{person}{Yehuda Koren},
  {and} \bibinfo{person}{Chris Volinsky}.} \bibinfo{year}{2008}\natexlab{}.
\newblock \showarticletitle{Collaborative filtering for implicit feedback
  datasets}. In \bibinfo{booktitle}{\emph{2008 Eighth IEEE International
  Conference on Data Mining}}. Ieee, \bibinfo{pages}{263--272}.
\newblock


\bibitem[\protect\citeauthoryear{Huang, Zhang, et~al\mbox{.}}{Huang
  et~al\mbox{.}}{2010}]%
        {huang2010benefit}
\bibfield{author}{\bibinfo{person}{Junzhou Huang}, \bibinfo{person}{Tong
  Zhang}, {et~al\mbox{.}}} \bibinfo{year}{2010}\natexlab{}.
\newblock \showarticletitle{The benefit of group sparsity}.
\newblock \bibinfo{journal}{\emph{The Annals of Statistics}}
  \bibinfo{volume}{38}, \bibinfo{number}{4} (\bibinfo{year}{2010}),
  \bibinfo{pages}{1978--2004}.
\newblock


\bibitem[\protect\citeauthoryear{Huang, Zhang, and Metaxas}{Huang
  et~al\mbox{.}}{2011}]%
        {huang2011learning}
\bibfield{author}{\bibinfo{person}{Junzhou Huang}, \bibinfo{person}{Tong
  Zhang}, {and} \bibinfo{person}{Dimitris Metaxas}.}
  \bibinfo{year}{2011}\natexlab{}.
\newblock \showarticletitle{Learning with structured sparsity}.
\newblock \bibinfo{journal}{\emph{Journal of Machine Learning Research}}
  \bibinfo{volume}{12}, \bibinfo{number}{Nov} (\bibinfo{year}{2011}),
  \bibinfo{pages}{3371--3412}.
\newblock


\bibitem[\protect\citeauthoryear{Jacobs, Jordan, Nowlan, and Hinton}{Jacobs
  et~al\mbox{.}}{1991}]%
        {jacobs1991adaptive}
\bibfield{author}{\bibinfo{person}{Robert~A Jacobs}, \bibinfo{person}{Michael~I
  Jordan}, \bibinfo{person}{Steven~J Nowlan}, {and} \bibinfo{person}{Geoffrey~E
  Hinton}.} \bibinfo{year}{1991}\natexlab{}.
\newblock \showarticletitle{Adaptive mixtures of local experts}.
\newblock \bibinfo{journal}{\emph{Neural computation}} \bibinfo{volume}{3},
  \bibinfo{number}{1} (\bibinfo{year}{1991}), \bibinfo{pages}{79--87}.
\newblock


\bibitem[\protect\citeauthoryear{Jordan and Jacobs}{Jordan and Jacobs}{1994}]%
        {jordan1994hierarchical}
\bibfield{author}{\bibinfo{person}{Michael~I Jordan} {and}
  \bibinfo{person}{Robert~A Jacobs}.} \bibinfo{year}{1994}\natexlab{}.
\newblock \showarticletitle{Hierarchical mixtures of experts and the EM
  algorithm}.
\newblock \bibinfo{journal}{\emph{Neural computation}} \bibinfo{volume}{6},
  \bibinfo{number}{2} (\bibinfo{year}{1994}), \bibinfo{pages}{181--214}.
\newblock


\bibitem[\protect\citeauthoryear{Kim, Wattenberg, Gilmer, Cai, Wexler, Viegas,
  et~al\mbox{.}}{Kim et~al\mbox{.}}{2018}]%
        {kim2018interpretability}
\bibfield{author}{\bibinfo{person}{Been Kim}, \bibinfo{person}{Martin
  Wattenberg}, \bibinfo{person}{Justin Gilmer}, \bibinfo{person}{Carrie Cai},
  \bibinfo{person}{James Wexler}, \bibinfo{person}{Fernanda Viegas},
  {et~al\mbox{.}}} \bibinfo{year}{2018}\natexlab{}.
\newblock \showarticletitle{Interpretability beyond feature attribution:
  Quantitative testing with concept activation vectors (tcav)}. In
  \bibinfo{booktitle}{\emph{International conference on machine learning}}.
  \bibinfo{pages}{2668--2677}.
\newblock


\bibitem[\protect\citeauthoryear{Kingma and Ba}{Kingma and Ba}{2014}]%
        {kingma2014adam}
\bibfield{author}{\bibinfo{person}{Diederik~P Kingma} {and}
  \bibinfo{person}{Jimmy Ba}.} \bibinfo{year}{2014}\natexlab{}.
\newblock \showarticletitle{Adam: A method for stochastic optimization}.
\newblock \bibinfo{journal}{\emph{Proceedings of the 3rd International
  Conference on Learning Representations}} (\bibinfo{year}{2014}).
\newblock


\bibitem[\protect\citeauthoryear{Li and She}{Li and She}{2017}]%
        {li2017collaborative}
\bibfield{author}{\bibinfo{person}{Xiaopeng Li} {and} \bibinfo{person}{James
  She}.} \bibinfo{year}{2017}\natexlab{}.
\newblock \showarticletitle{Collaborative variational autoencoder for
  recommender systems}. In \bibinfo{booktitle}{\emph{Proceedings of the 23rd
  ACM SIGKDD international conference on knowledge discovery and data mining}}.
  \bibinfo{pages}{305--314}.
\newblock


\bibitem[\protect\citeauthoryear{Liang, Krishnan, Hoffman, and Jebara}{Liang
  et~al\mbox{.}}{2018}]%
        {liang2018variational}
\bibfield{author}{\bibinfo{person}{Dawen Liang}, \bibinfo{person}{Rahul~G
  Krishnan}, \bibinfo{person}{Matthew~D Hoffman}, {and} \bibinfo{person}{Tony
  Jebara}.} \bibinfo{year}{2018}\natexlab{}.
\newblock \showarticletitle{Variational autoencoders for collaborative
  filtering}. In \bibinfo{booktitle}{\emph{Proceedings of the 2018 World Wide
  Web Conference}}. \bibinfo{pages}{689--698}.
\newblock


\bibitem[\protect\citeauthoryear{Lou, Caruana, Gehrke, and Hooker}{Lou
  et~al\mbox{.}}{2013}]%
        {lou2013accurate}
\bibfield{author}{\bibinfo{person}{Yin Lou}, \bibinfo{person}{Rich Caruana},
  \bibinfo{person}{Johannes Gehrke}, {and} \bibinfo{person}{Giles Hooker}.}
  \bibinfo{year}{2013}\natexlab{}.
\newblock \showarticletitle{Accurate intelligible models with pairwise
  interactions}. In \bibinfo{booktitle}{\emph{Proceedings of the 19th ACM
  SIGKDD international conference on Knowledge discovery and data mining}}.
  \bibinfo{pages}{623--631}.
\newblock


\bibitem[\protect\citeauthoryear{Lundberg and Lee}{Lundberg and Lee}{2017}]%
        {NIPS2017_7062}
\bibfield{author}{\bibinfo{person}{Scott~M Lundberg} {and}
  \bibinfo{person}{Su-In Lee}.} \bibinfo{year}{2017}\natexlab{}.
\newblock \showarticletitle{A Unified Approach to Interpreting Model
  Predictions}.
\newblock In \bibinfo{booktitle}{\emph{Advances in Neural Information
  Processing Systems 30}}, \bibfield{editor}{\bibinfo{person}{I.~Guyon},
  \bibinfo{person}{U.~V. Luxburg}, \bibinfo{person}{S.~Bengio},
  \bibinfo{person}{H.~Wallach}, \bibinfo{person}{R.~Fergus},
  \bibinfo{person}{S.~Vishwanathan}, {and} \bibinfo{person}{R.~Garnett}}
  (Eds.). \bibinfo{publisher}{Curran Associates, Inc.},
  \bibinfo{pages}{4765--4774}.
\newblock
\urldef\tempurl%
\url{http://papers.nips.cc/paper/7062-a-unified-approach-to-interpreting-model-predictions.pdf}
\showURL{%
\tempurl}


\bibitem[\protect\citeauthoryear{Maziarka, Danel, Mucha, Rataj, Tabor, and
  Jastrzębski}{Maziarka et~al\mbox{.}}{2020}]%
        {maziarka2020molecule}
\bibfield{author}{\bibinfo{person}{Łukasz Maziarka}, \bibinfo{person}{Tomasz
  Danel}, \bibinfo{person}{Sławomir Mucha}, \bibinfo{person}{Krzysztof Rataj},
  \bibinfo{person}{Jacek Tabor}, {and} \bibinfo{person}{Stanisław
  Jastrzębski}.} \bibinfo{year}{2020}\natexlab{}.
\newblock \showarticletitle{Molecule Attention Transformer}.
\newblock \bibinfo{journal}{\emph{arXiv preprint arXiv:2002.08264}}
  (\bibinfo{year}{2020}).
\newblock


\bibitem[\protect\citeauthoryear{Redmon, Divvala, Girshick, and Farhadi}{Redmon
  et~al\mbox{.}}{2016}]%
        {redmon2016you}
\bibfield{author}{\bibinfo{person}{Joseph Redmon}, \bibinfo{person}{Santosh
  Divvala}, \bibinfo{person}{Ross Girshick}, {and} \bibinfo{person}{Ali
  Farhadi}.} \bibinfo{year}{2016}\natexlab{}.
\newblock \showarticletitle{You only look once: Unified, real-time object
  detection}. In \bibinfo{booktitle}{\emph{Proceedings of the IEEE conference
  on computer vision and pattern recognition}}. \bibinfo{pages}{779--788}.
\newblock


\bibitem[\protect\citeauthoryear{Ribeiro, Singh, and Guestrin}{Ribeiro
  et~al\mbox{.}}{2016}]%
        {ribeiro2016should}
\bibfield{author}{\bibinfo{person}{Marco~Tulio Ribeiro},
  \bibinfo{person}{Sameer Singh}, {and} \bibinfo{person}{Carlos Guestrin}.}
  \bibinfo{year}{2016}\natexlab{}.
\newblock \showarticletitle{" Why should i trust you?" Explaining the
  predictions of any classifier}. In \bibinfo{booktitle}{\emph{Proceedings of
  the 22nd ACM SIGKDD international conference on knowledge discovery and data
  mining}}. \bibinfo{pages}{1135--1144}.
\newblock


\bibitem[\protect\citeauthoryear{Rudin}{Rudin}{2019}]%
        {rudin2019stop}
\bibfield{author}{\bibinfo{person}{Cynthia Rudin}.}
  \bibinfo{year}{2019}\natexlab{}.
\newblock \showarticletitle{Stop explaining black box machine learning models
  for high stakes decisions and use interpretable models instead}.
\newblock \bibinfo{journal}{\emph{Nature Machine Intelligence}}
  \bibinfo{volume}{1}, \bibinfo{number}{5} (\bibinfo{year}{2019}),
  \bibinfo{pages}{206--215}.
\newblock


\bibitem[\protect\citeauthoryear{Schulz, Sixt, Tombari, and Landgraf}{Schulz
  et~al\mbox{.}}{2019}]%
        {schulz2020restricting}
\bibfield{author}{\bibinfo{person}{Karl Schulz}, \bibinfo{person}{Leon Sixt},
  \bibinfo{person}{Federico Tombari}, {and} \bibinfo{person}{Tim Landgraf}.}
  \bibinfo{year}{2019}\natexlab{}.
\newblock \showarticletitle{Restricting the Flow: Information Bottlenecks for
  Attribution}. In \bibinfo{booktitle}{\emph{International Conference on
  Learning Representations}}.
\newblock


\bibitem[\protect\citeauthoryear{Selvaraju, Cogswell, Das, Vedantam, Parikh,
  and Batra}{Selvaraju et~al\mbox{.}}{2017}]%
        {selvaraju2017grad}
\bibfield{author}{\bibinfo{person}{Ramprasaath~R Selvaraju},
  \bibinfo{person}{Michael Cogswell}, \bibinfo{person}{Abhishek Das},
  \bibinfo{person}{Ramakrishna Vedantam}, \bibinfo{person}{Devi Parikh}, {and}
  \bibinfo{person}{Dhruv Batra}.} \bibinfo{year}{2017}\natexlab{}.
\newblock \showarticletitle{Grad-cam: Visual explanations from deep networks
  via gradient-based localization}. In \bibinfo{booktitle}{\emph{Proceedings of
  the IEEE international conference on computer vision}}.
  \bibinfo{pages}{618--626}.
\newblock


\bibitem[\protect\citeauthoryear{Shenbin, Alekseev, Tutubalina, Malykh, and
  Nikolenko}{Shenbin et~al\mbox{.}}{2020}]%
        {shenbin2020recvae}
\bibfield{author}{\bibinfo{person}{Ilya Shenbin}, \bibinfo{person}{Anton
  Alekseev}, \bibinfo{person}{Elena Tutubalina}, \bibinfo{person}{Valentin
  Malykh}, {and} \bibinfo{person}{Sergey~I Nikolenko}.}
  \bibinfo{year}{2020}\natexlab{}.
\newblock \showarticletitle{RecVAE: A New Variational Autoencoder for Top-N
  Recommendations with Implicit Feedback}. In
  \bibinfo{booktitle}{\emph{Proceedings of the 13th International Conference on
  Web Search and Data Mining}}. \bibinfo{pages}{528--536}.
\newblock


\bibitem[\protect\citeauthoryear{Shrikumar, Greenside, and Kundaje}{Shrikumar
  et~al\mbox{.}}{2017}]%
        {shrikumar2017learning}
\bibfield{author}{\bibinfo{person}{Avanti Shrikumar}, \bibinfo{person}{Peyton
  Greenside}, {and} \bibinfo{person}{Anshul Kundaje}.}
  \bibinfo{year}{2017}\natexlab{}.
\newblock \showarticletitle{Learning Important Features Through Propagating
  Activation Differences}. In \bibinfo{booktitle}{\emph{International
  Conference on Machine Learning}}. \bibinfo{pages}{3145--3153}.
\newblock


\bibitem[\protect\citeauthoryear{Simonyan, Vedaldi, and Zisserman}{Simonyan
  et~al\mbox{.}}{2014}]%
        {simonyan2013deep}
\bibfield{author}{\bibinfo{person}{Karen Simonyan}, \bibinfo{person}{Andrea
  Vedaldi}, {and} \bibinfo{person}{Andrew Zisserman}.}
  \bibinfo{year}{2014}\natexlab{}.
\newblock \showarticletitle{Deep inside convolutional networks: Visualising
  image classification models and saliency maps}.
\newblock \bibinfo{journal}{\emph{Workshop, ICLR}} (\bibinfo{year}{2014}).
\newblock


\bibitem[\protect\citeauthoryear{Sinha and Swearingen}{Sinha and
  Swearingen}{2002}]%
        {sinha2002role}
\bibfield{author}{\bibinfo{person}{Rashmi Sinha} {and} \bibinfo{person}{Kirsten
  Swearingen}.} \bibinfo{year}{2002}\natexlab{}.
\newblock \showarticletitle{The role of transparency in recommender systems}.
  In \bibinfo{booktitle}{\emph{CHI'02 extended abstracts on Human factors in
  computing systems}}. \bibinfo{pages}{830--831}.
\newblock


\bibitem[\protect\citeauthoryear{Smilkov, Thorat, Kim, Vi{\'e}gas, and
  Wattenberg}{Smilkov et~al\mbox{.}}{2017}]%
        {smilkov2017smoothgrad}
\bibfield{author}{\bibinfo{person}{Daniel Smilkov}, \bibinfo{person}{Nikhil
  Thorat}, \bibinfo{person}{Been Kim}, \bibinfo{person}{Fernanda Vi{\'e}gas},
  {and} \bibinfo{person}{Martin Wattenberg}.} \bibinfo{year}{2017}\natexlab{}.
\newblock \showarticletitle{Smoothgrad: removing noise by adding noise}.
\newblock \bibinfo{journal}{\emph{Workshop on Visualization for Deep Learning,
  ICML}} (\bibinfo{year}{2017}).
\newblock


\bibitem[\protect\citeauthoryear{Tintarev and Masthoff}{Tintarev and
  Masthoff}{2007}]%
        {tintarev2007survey}
\bibfield{author}{\bibinfo{person}{Nava Tintarev} {and} \bibinfo{person}{Judith
  Masthoff}.} \bibinfo{year}{2007}\natexlab{}.
\newblock \showarticletitle{A survey of explanations in recommender systems}.
  In \bibinfo{booktitle}{\emph{2007 IEEE 23rd international conference on data
  engineering workshop}}. IEEE, \bibinfo{pages}{801--810}.
\newblock


\bibitem[\protect\citeauthoryear{Tishby, Pereira, and Bialek}{Tishby
  et~al\mbox{.}}{1999}]%
        {tishby2000information}
\bibfield{author}{\bibinfo{person}{Naftali Tishby}, \bibinfo{person}{Fernando~C
  Pereira}, {and} \bibinfo{person}{William Bialek}.}
  \bibinfo{year}{1999}\natexlab{}.
\newblock \showarticletitle{The information bottleneck method}.
\newblock \bibinfo{journal}{\emph{The 37th annual Allerton Conf. on
  Communication, Control, and Computing}} (\bibinfo{year}{1999}),
  \bibinfo{pages}{368--377}.
\newblock


\bibitem[\protect\citeauthoryear{Van~den Oord, Dieleman, and Schrauwen}{Van~den
  Oord et~al\mbox{.}}{2013}]%
        {van2013deep}
\bibfield{author}{\bibinfo{person}{Aaron Van~den Oord}, \bibinfo{person}{Sander
  Dieleman}, {and} \bibinfo{person}{Benjamin Schrauwen}.}
  \bibinfo{year}{2013}\natexlab{}.
\newblock \showarticletitle{Deep content-based music recommendation}. In
  \bibinfo{booktitle}{\emph{Advances in neural information processing
  systems}}. \bibinfo{pages}{2643--2651}.
\newblock


\bibitem[\protect\citeauthoryear{Vaswani, Shazeer, Parmar, Uszkoreit, Jones,
  Gomez, Kaiser, and Polosukhin}{Vaswani et~al\mbox{.}}{2017}]%
        {vaswani2017attention}
\bibfield{author}{\bibinfo{person}{Ashish Vaswani}, \bibinfo{person}{Noam
  Shazeer}, \bibinfo{person}{Niki Parmar}, \bibinfo{person}{Jakob Uszkoreit},
  \bibinfo{person}{Llion Jones}, \bibinfo{person}{Aidan~N Gomez},
  \bibinfo{person}{{\L}ukasz Kaiser}, {and} \bibinfo{person}{Illia
  Polosukhin}.} \bibinfo{year}{2017}\natexlab{}.
\newblock \showarticletitle{Attention is all you need}. In
  \bibinfo{booktitle}{\emph{Advances in neural information processing
  systems}}. \bibinfo{pages}{5998--6008}.
\newblock


\bibitem[\protect\citeauthoryear{Voita, Talbot, Moiseev, Sennrich, and
  Titov}{Voita et~al\mbox{.}}{2019}]%
        {voita2019analyzing}
\bibfield{author}{\bibinfo{person}{Elena Voita}, \bibinfo{person}{David
  Talbot}, \bibinfo{person}{Fedor Moiseev}, \bibinfo{person}{Rico Sennrich},
  {and} \bibinfo{person}{Ivan Titov}.} \bibinfo{year}{2019}\natexlab{}.
\newblock \showarticletitle{Analyzing Multi-Head Self-Attention: Specialized
  Heads Do the Heavy Lifting, the Rest Can Be Pruned}. In
  \bibinfo{booktitle}{\emph{Proceedings of the 57th Annual Meeting of the
  Association for Computational Linguistics}}. \bibinfo{pages}{5797--5808}.
\newblock


\bibitem[\protect\citeauthoryear{Voynov and Babenko}{Voynov and
  Babenko}{2019}]%
        {voynov2019rpgan}
\bibfield{author}{\bibinfo{person}{Andrey Voynov} {and} \bibinfo{person}{Artem
  Babenko}.} \bibinfo{year}{2019}\natexlab{}.
\newblock \showarticletitle{RPGAN: GANs Interpretability via Random Routing}.
\newblock \bibinfo{journal}{\emph{arXiv preprint arXiv:1912.10920}}
  (\bibinfo{year}{2019}).
\newblock


\bibitem[\protect\citeauthoryear{Zhao, Rocha, Yu, et~al\mbox{.}}{Zhao
  et~al\mbox{.}}{2009}]%
        {zhao2009composite}
\bibfield{author}{\bibinfo{person}{Peng Zhao}, \bibinfo{person}{Guilherme
  Rocha}, \bibinfo{person}{Bin Yu}, {et~al\mbox{.}}}
  \bibinfo{year}{2009}\natexlab{}.
\newblock \showarticletitle{The composite absolute penalties family for grouped
  and hierarchical variable selection}.
\newblock \bibinfo{journal}{\emph{The Annals of Statistics}}
  \bibinfo{volume}{37}, \bibinfo{number}{6A} (\bibinfo{year}{2009}),
  \bibinfo{pages}{3468--3497}.
\newblock


\bibitem[\protect\citeauthoryear{Zhou, Khosla, Lapedriza, Oliva, and
  Torralba}{Zhou et~al\mbox{.}}{2016}]%
        {zhou2016learning}
\bibfield{author}{\bibinfo{person}{Bolei Zhou}, \bibinfo{person}{Aditya
  Khosla}, \bibinfo{person}{Agata Lapedriza}, \bibinfo{person}{Aude Oliva},
  {and} \bibinfo{person}{Antonio Torralba}.} \bibinfo{year}{2016}\natexlab{}.
\newblock \showarticletitle{Learning deep features for discriminative
  localization}. In \bibinfo{booktitle}{\emph{Proceedings of the IEEE
  conference on computer vision and pattern recognition}}.
  \bibinfo{pages}{2921--2929}.
\newblock


\bibitem[\protect\citeauthoryear{Zhu and Chen}{Zhu and Chen}{2019}]%
        {zhu2019session}
\bibfield{author}{\bibinfo{person}{Lin Zhu} {and} \bibinfo{person}{Yihong
  Chen}.} \bibinfo{year}{2019}\natexlab{}.
\newblock \showarticletitle{Session-based Sequential Skip Prediction via
  Recurrent Neural Networks}.
\newblock \bibinfo{journal}{\emph{arXiv preprint arXiv:1902.04743}}
  (\bibinfo{year}{2019}).
\newblock


\bibitem[\protect\citeauthoryear{Zilke, Menc{\'\i}a, and Janssen}{Zilke
  et~al\mbox{.}}{2016}]%
        {zilke2016deepred}
\bibfield{author}{\bibinfo{person}{Jan~Ruben Zilke},
  \bibinfo{person}{Eneldo~Loza Menc{\'\i}a}, {and} \bibinfo{person}{Frederik
  Janssen}.} \bibinfo{year}{2016}\natexlab{}.
\newblock \showarticletitle{Deepred--rule extraction from deep neural
  networks}. In \bibinfo{booktitle}{\emph{International Conference on Discovery
  Science}}. Springer, \bibinfo{pages}{457--473}.
\newblock


\end{thebibliography}

\end{document}